\documentclass[letterpaper]{article} % DO NOT CHANGE THIS

\usepackage{aaai2026}  % DO NOT CHANGE THIS
\usepackage{times}  % DO NOT CHANGE THIS
\usepackage{helvet}  % DO NOT CHANGE THIS
\usepackage{courier}  % DO NOT CHANGE THIS
\usepackage[hyphens]{url}  % DO NOT CHANGE THIS
\usepackage{graphicx} % DO NOT CHANGE THIS
\usepackage{amsmath}  %
\usepackage{booktabs} % 三线表必备
\usepackage{array}    % 对齐控制
\usepackage{amsmath}  % 数学符号
\usepackage{xcolor}   % 颜色控制（用于强调）
\usepackage{graphicx} % 调整表格大小
\usepackage{cuted} 
\usepackage{multirow}
\usepackage{afterpage}
\usepackage{amssymb}
\newcommand{\firstplace}[1]{\textbf{#1}}

\newcommand{\secondplace}[1]{\underline{#1}}

\def\UrlFont{\rm}  % DO NOT CHANGE THIS
\usepackage{natbib}  % DO NOT CHANGE THIS AND DO NOT ADD ANY OPTIONS TO IT
\usepackage{caption} % DO NOT CHANGE THIS AND DO NOT ADD ANY OPTIONS TO IT
\usepackage{algorithm}
\usepackage{algorithmic}

\usepackage{newfloat}
\usepackage{listings}
\DeclareCaptionStyle{ruled}{labelfont=normalfont,labelsep=colon,strut=off} % DO NOT CHANGE THIS
\floatstyle{ruled}
\newfloat{listing}{tb}{lst}{}
\floatname{listing}{Listing}
\title{FreqCycle: A Multi-Scale Time-Frequency Analysis Method for Time Series Forecasting}
\author{
    Boya Zhang,
    Shuaijie Yin,
    Huiwen Zhu,
    Xing He\thanks{Corresponding Author}
}
\affiliations{
    Shanghai Jiao Tong University, Shanghai, China \\
    \{zby112200, yinshuaijie, shu\_yuan, xhe\}@sjtu.edu.cn
}

\begin{document}

\maketitle

\begin{abstract}
Mining time-frequency features is critical for time series forecasting. Existing research has predominantly focused on modeling low-frequency patterns, where most time series energy is concentrated. The overlooking of mid to high frequency continues to limit further performance gains in deep learning models. We propose \textbf{FreqCycle}, a novel framework integrating: (i) a \textbf{Filter-Enhanced Cycle Forecasting (FECF)} module to extract low-frequency features by explicitly learning shared periodic patterns in the time domain, and (ii) a \textbf{Segmented Frequency-domain Pattern Learning (SFPL)} module to enhance mid to high frequency energy proportion via learnable filters and adaptive weighting. Furthermore, time series data often exhibit coupled multi-periodicity, such as intertwined weekly and daily cycles. To address coupled multi-periodicity as well as long lookback window challenges, we extend FreqCycle hierarchically into \textbf{MFreqCycle}, which decouples nested periodic features through cross-scale interactions. Extensive experiments on seven diverse domain benchmarks demonstrate that FreqCycle achieves state-of-the-art accuracy while maintaining faster inference speeds, striking an optimal balance between performance and efficiency.
\end{abstract}

\begin{links}
    \link{Code}https://github.com/boya-zhang-ai/FreqCycle
\end{links}

\section{Introduction}

Time series forecasting (TSF) plays a pivotal role in numerous domains, including energy optimization\cite{WOS:000701874500003}, weather forecasting\cite{WOS:000999552700001,fan2023dishtsgeneralparadigmalleviating}, and economic analysis\cite{ARIMA}, where it serves as a priori guidance for proactive planning and decision-making. The ubiquity and significance of TSF tasks have spurred a proliferation of methodological advances in this field.

Traditional forecasting approaches, such as ARIMA \cite{ARIMA} and LSTM \cite{LSTM}, were predominantly employed in early TSF research.

The rapid development of deep learning has given rise to a diverse array of neural network based forecasting models. Recurrent Neural Network, RNN-based methods (e.g., SegRNN \cite{SegRNN}, WITRAN \cite{jia2023witran}) are frequently hampered by gradient vanishing or explosion issues, which limit their ability to model long-term dependencies. Convolutional Neural Network, CNN-based approaches (e.g., TCN \cite{TCN}, SCINet \cite{SCINet}) leverage parallel computation but often exhibit restricted receptive fields, potentially overlooking global temporal relationships. Transformer-based techniques (e.g., Autoformer\cite{autoformer}, PatchTST\cite{PatchTST}, iTransformer\cite{iTransformer}) excel at modeling global dependencies through self-attention mechanisms, albeit at the cost of high computational complexity. Meanwhile, Multilayer Perceptron MLP-based models (e.g., DLinear\cite{DLinear}, CycleNet\cite{cyclenet}) offer simplicity and training efficiency, demonstrating notable efficacy in low-frequency component prediction.
\begin{figure}[tb]
\centering
\includegraphics[width=1\columnwidth]{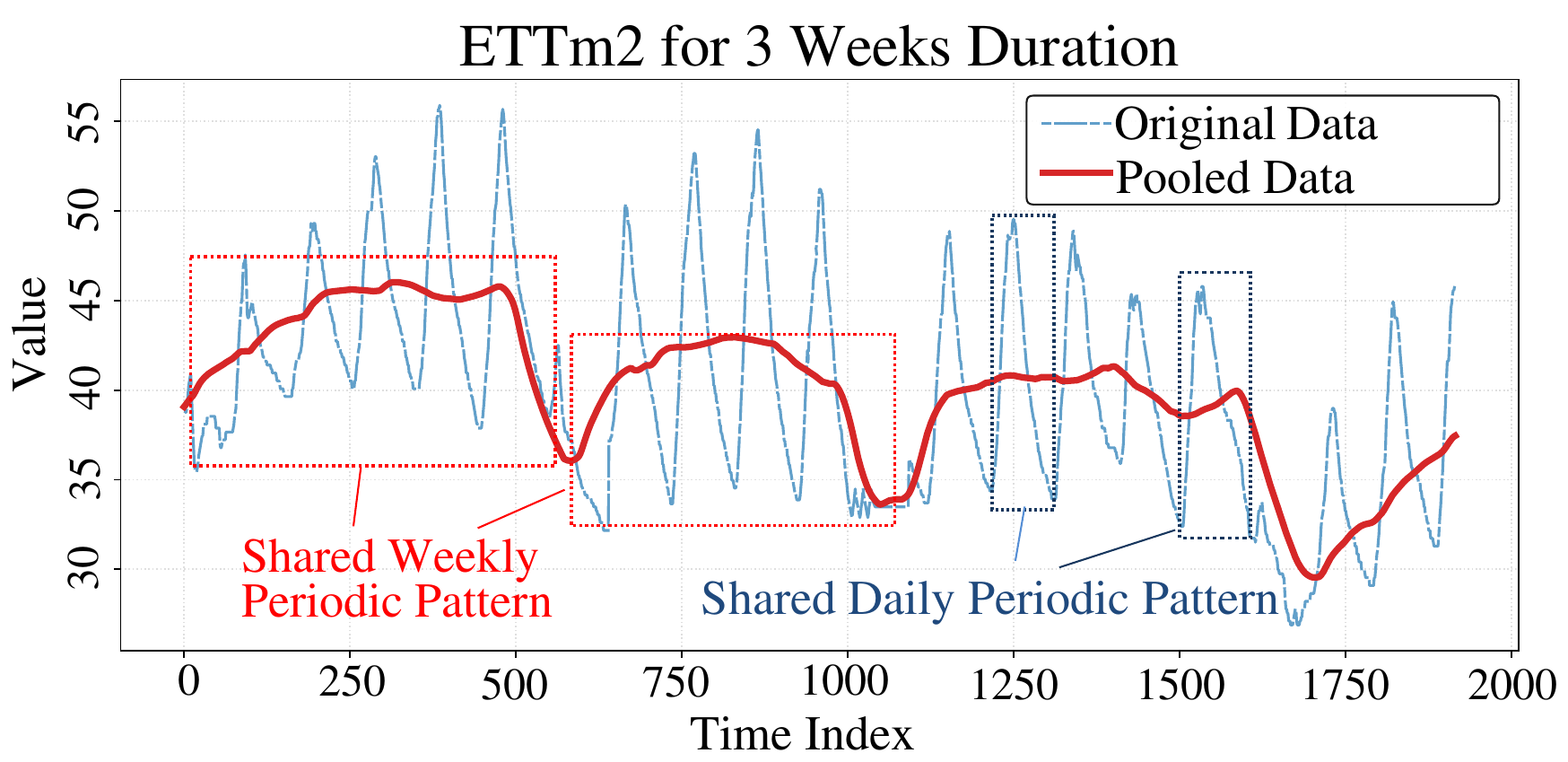}
\caption{Shared periodic patterns present in the ETTm2 dataset.}
\label{fig5}
\end{figure}
\begin{figure}[h]
\centering
\includegraphics[width=1\columnwidth]{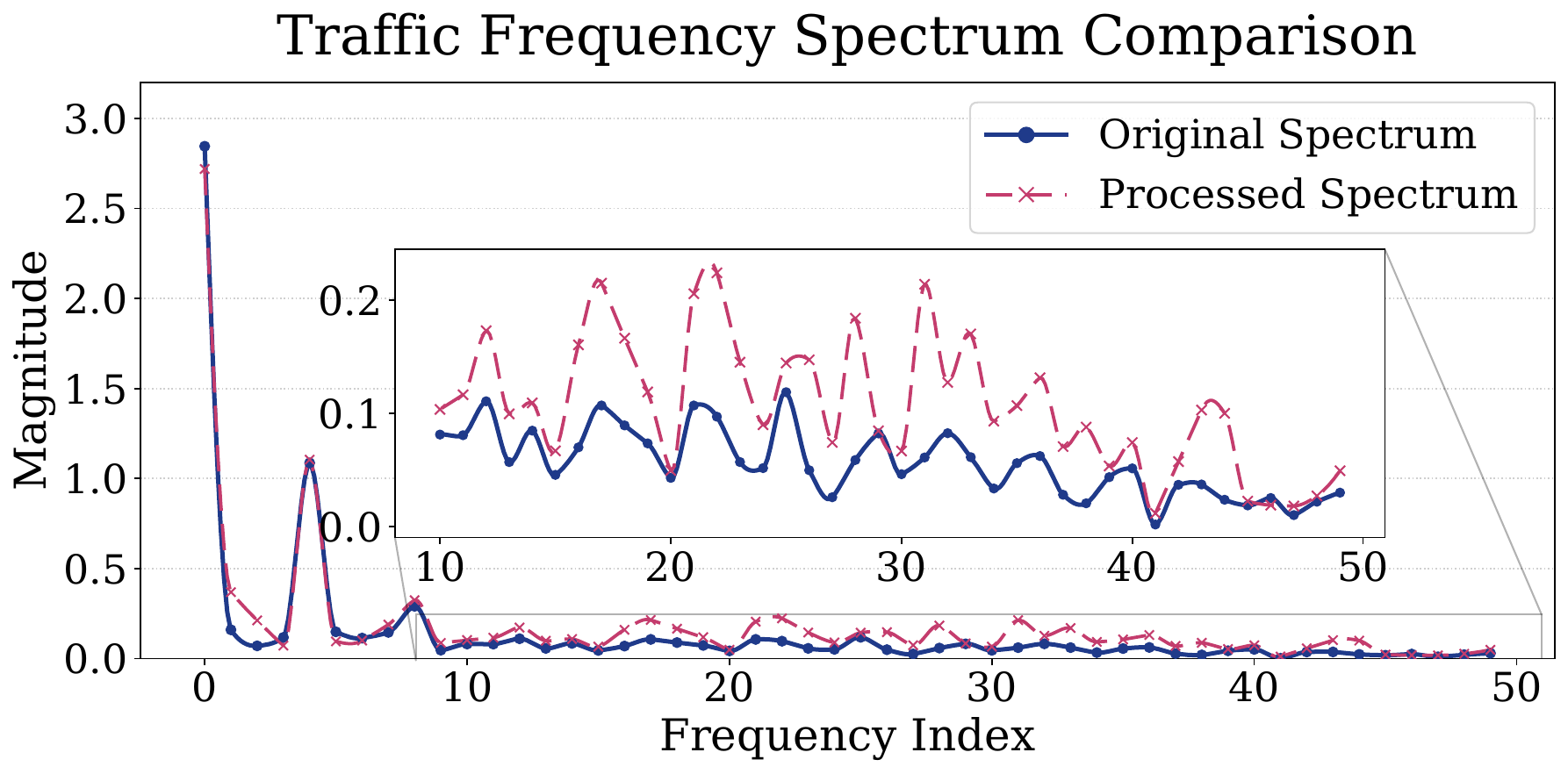}
\caption{Frequency spectrum of Traffic dataset before and after SFPL enhancement}
\label{fig6}
\end{figure}

MLP-based methods have garnered significant attention due to their low complexity and competitive performance. However, current MLP-based approaches exhibit several limitations: (i) Deep learning models aim to construct complex architectures to extract periodic patterns from long-range dependencies. However, we opt for a more direct and effective approach by explicitly learning shared daily and weekly periodic patterns through parametric modeling. Figure \ref{fig5} shows the daily periodic patterns in the ETT dataset and the observable weekly patterns after average pooling. (ii) From a frequency-domain perspective, although simple models (e.g., DLinear \cite{DLinear}) excel at learning low-frequency periodic components, their fundamental architecture—built upon Linear Layers or MLPs—inherently limits their ability to effectively represent mid to high frequency components, which capture short-term fluctuations and aperiodic features. As shown in Figure \ref{fig6}, these mid to high frequency components contribute very little energy proportion, urgently requiring effective methods to enhance the energy contribution of these critical frequency components. (iii) Existing architectures are not specifically designed to address the multi-scale nested periodicity (such as coupled daily-weekly cycles) inherent in time series data. Naively employing MLP-based models could result in compromised performance or reduced efficiency, thereby demanding the construction of an improved framework that can adaptively model multi-level periodic structures in a coordinated manner.

To address the challenges in time series forecasting, this paper makes the following key contributions:
\begin{itemize}
\item We introduce Filter-Enhanced Cycle Forecasting (FECF), a method that explicitly learns shared periodic patterns, avoiding complex architectures to learn long-range dependencies.
\item We propose the Segmented Frequency Pattern Learning (SFPL) method, which enhances the energy contribution of mid to high frequency components. This approach effectively resolves the long-standing limitation of insufficient key frequency modeling in conventional methods, improving prediction accuracy for aperiodic elements.
\item To address the critical challenge of coupled multi-periodicity under long lookback windows, we developed MFreqCycle - an architecture that decouples and jointly models nested periodic features through a multi-scale approach.
\item We conduct extensive experiments on seven time series forecasting benchmarks. The results demonstrate that our model achieves superior performance in both effectiveness and efficiency.
\end{itemize}

\section{Related Work}
\subsection{MLP-based Time Series Model}
Several studies have investigated the application of MLP-based deep learning networks for time series forecasting. N-BEATS \cite{N-BEATS} employs stacked MLP layers with dual residual learning to process input data and generate iterative future predictions. LightTS \cite{lightTS} adopts a lightweight sampling-oriented MLP architecture to reduce complexity and computational overhead while maintaining prediction accuracy. DLinear \cite{DLinear} introduces a set of simple single-layer Linear models to capture temporal relationships between input and output sequences. SparseTSF \cite{sparsetsf} utilizes cross-cycle sparse prediction techniques to decouple periodic and trend components. CycleNet \cite{cyclenet} demonstrates remarkable performance in day-ahead forecasting by explicitly modeling periodic patterns through an integrated framework combining cyclical decomposition and MLP-based residual prediction. TimeMixer \cite{timemixer} introduces a novel architecture featuring Past-Decomposable-Mixing for multiscale representation learning and Future-Multipredictor-Mixing to ensemble multi-scale forecasting skills. These studies collectively validate the effectiveness of MLP-based architectures in time series forecasting tasks, motivating our further development of MLP-based deep learning prediction algorithms in this work.

\subsection{Time Series Modeling with Frequency Learning}
Recent years have witnessed increasing integration of frequency techniques into deep learning models for time series forecasting, capitalizing on their computational efficiency and energy compaction properties \cite{yi2023survey}. Specifically, Autoformer \cite{autoformer} replaces self-attention with an autocorrelation mechanism incorporating Fast Fourier Transform (FFT), while FEDformer \cite{fedformer} proposes a DFT-based frequency-enhanced attention that computes attention weights through query-key spectral analysis. FiLM \cite{zhou2022film} employs Fourier analysis to preserve historical patterns while filtering noise, and FreTS \cite{yi2023frets} introduces frequency-domain MLPs to learn cross-channel and temporal dependencies. FITS \cite{xufits} applies low-pass filtering before performing complex-valued Linear transformations in the frequency domain. FilterNet \cite{yi2024filternet} introduces plain shaping filters for baseline frequency processing and contextual shaping filters for adaptive spectrum modulation. Notably, most existing frequency-based methods primarily focus on low-frequency components through conventional or adaptive filtering. In contrast, our work proposes a novel frequency learning architecture that specifically enhances mid to high frequency energy representation.
\begin{figure*}[h]
\centering
\includegraphics[width=2.1\columnwidth]{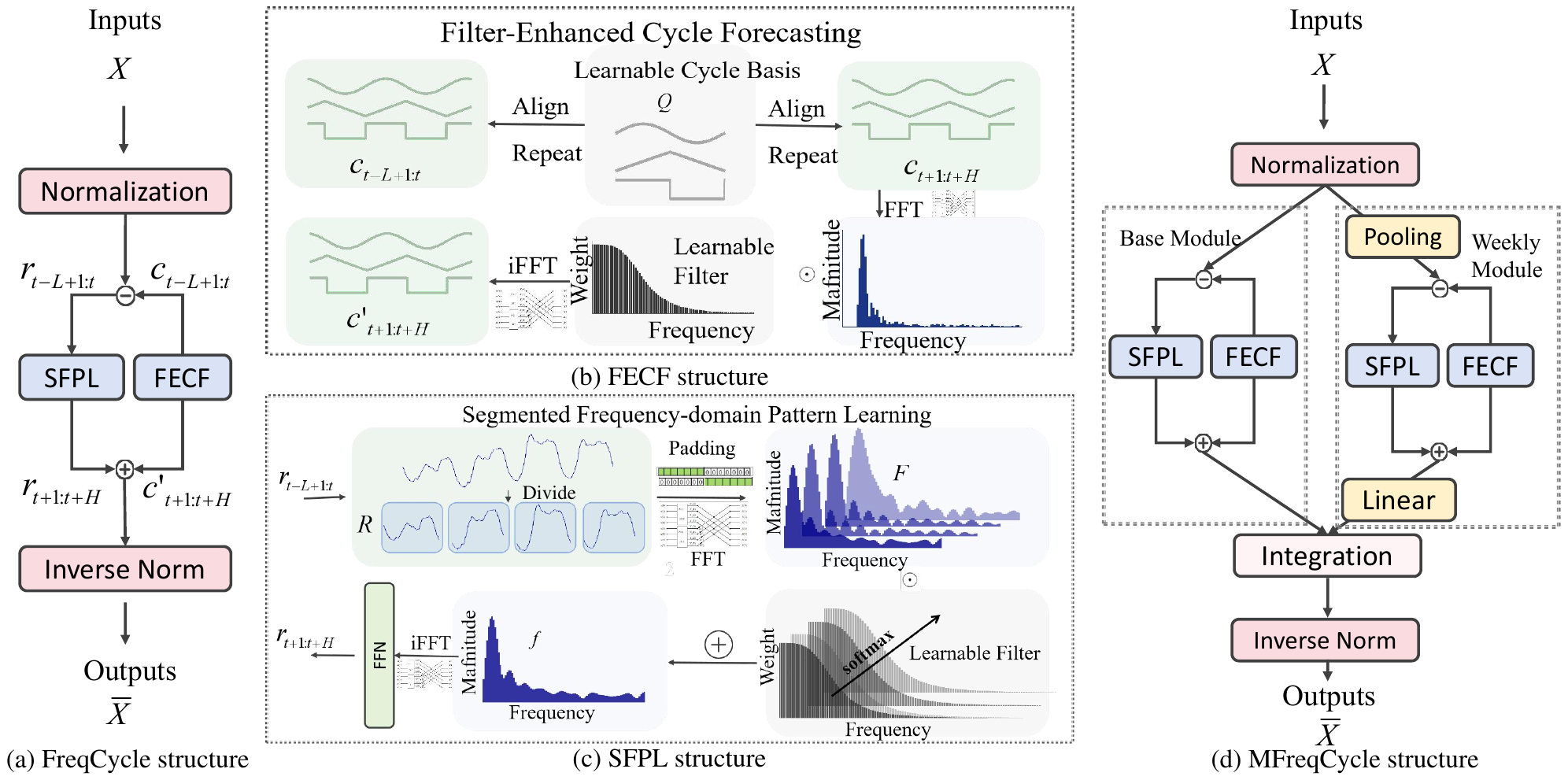} % Reduce the figure size so that it is slightly narrower than the column. Don't use precise values for figure width.This setup will avoid overfull boxes.
\caption{The architecture of FreqCycle. (a) The overall architecture of FreqCycle. (b) FECF block is employed to explicitly learns shared periodic patterns in the time domain; (c) SFPL extracts critical frequency information in the spectral domain via learnable filters and adaptive weighting.(d) MFreqCycle explicitly decouples and models nested periodic features through cross-scale interactions.}
\label{fig0}
\end{figure*}
\section{Methodology}
In this section, we first formalize the TSF problem and introduce the time-frequency transformation theory along with its inherent time-frequency duality, thereby establishing the theoretical foundation for our methodology. We then present the two core components of FreqCycle: (i) the Filter-Enhanced Cycle Forecasting (FECF) technique, and (ii) the Segmented Frequency-domain Pattern Learning (SFPL) module. To address the critical challenge of coupled periodic patterns in real-world time series, we further develop MFreqCycle by incorporating cross-scale feature fusion mechanisms into the FreqCycle architecture, enabling decoupling and joint modeling of nested periodic features. The overall structure is illustrated in Figure \ref{fig0}.
\subsection{Preliminaries}\label{AA}
\noindent \hspace{1em}\textup{1) Problem Definition:} Given a time series \( X \in \mathbb{R}^{L \times D} \) with \( L \) historical observations and \( D \) variables, the goal is to predict future values \( \overline{X} \in \mathbb{R}^{H \times D} \) where \( H \) is the forecast horizon. The mapping function can be formalized as:
\[ f :x_{t-L+1:t}\in \mathbb{R}^{L \times D} \rightarrow \overline{x}_{t+1:t+H}\in\mathbb{R}^{H \times D} \]\label{eq}
\noindent \hspace{1em}\textup{2) Discrete Fourier Transform:} In TSF problems, time series data are discrete samples that vary over time. The Discrete Fourier Transform (DFT) is a widely used method for converting discrete data from the time domain to the frequency domain. For a discrete sequence x[n] of length N, its DFT is defined as:
\begin{equation}
X[k] = \sum_{n=0}^{N-1} x[n] e^{-j\frac{2\pi}{N}kn}, \quad k = 0, 1, \ldots, N-1  
\end{equation}
where \(j = \sqrt{-1}\) represents the imaginary unit.

The Inverse Discrete Fourier Transform (iDFT) is the reverse operation of DFT, which reconstructs the time-domain signal from its frequency-domain representation:
\begin{equation}
x[n] = \frac{1}{N} \sum_{k=0}^{N-1} X[k] e^{j \frac{2\pi}{N} kn}, \quad n = 0, 1, \ldots, N - 1
\end{equation}

In practical applications, Fast Fourier Transform (FFT) and inverse Fast Fourier Transform (iFFT) are commonly employed instead of DFT and iDFT to significantly reduce computational complexity and improve operational efficiency. This efficient time-frequency transformation serves as the fundamental basis for implementing various frequency-domain operations on time series data.

\noindent \hspace{1em}\textup{3) Duality of Time and Frequency Domain:}
For two discrete-time sequences x[n] and h[n], the convolution operation in the time domain is computed as follows:
\begin{equation}
y[n] = (x \ast h)[n]= \sum_{m=0}^{N-1} x[m] h[n - m]
\end{equation}

The convolution operation in the time domain is equivalent to point-wise multiplication in the frequency domain:
\begin{equation}
\mathcal{F}\{ x \ast h \} = X[k] \cdot H[k]
\end{equation}

Conversely, multiplication in the time domain corresponds to convolution in the frequency domain.Based on the time-frequency duality property between convolution and multiplication, numerous researchers \cite{zhao2024disentangled,yi2024filternet} have adopted frequency-domain pointwise multiplication operations to replace time-domain circular convolution for enhanced computational efficiency.

Another industrially significant time-frequency duality manifests in the deterministic relationship between temporal sampling interval \(\Delta t\) and signal bandwidth \(f_{\max}\), as well as the inverse dependence of frequency resolution \(\Delta f\) on total observation duration \(T\). 

\subsection{Filter-Enhanced Cycle Forecasting}
The Filter-Enhanced Cyclical Forecasting (FECF) technique represents an improvement over the original Residual Cycle Forecasting framework\cite{cyclenet}, specifically designed to enhance the learning of low-frequency periodic patterns in time series data. 

Given input sequences with \(D\) channels and a predefined base period length \(W\), the method first generates a learnable cyclical basis \(Q\in \mathbb{R}^{W \times D}\) (initialized as a zero matrix) that is globally shared across the temporal dimension. Through periodic replication operations \cite{cyclenet}, this basis produces a cyclical component \(c_{t-L+1:t}\) matching the original sequence length \(L\), as well as a cyclical component \(c_{t+1:t+H}\) matching the predict length \(H\). To prioritize low-frequency information capture, the generated cyclical component \(c_{t+1:t+H}\) undergoes processing via adaptive filtering. Specifically, the adaptive filtering enhanced periodic learning module focuses on amplifying low-frequency information while attenuating mid- and high-frequency components that may degrade periodic pattern learning performance. The workflow is implemented as follows:
\begin{equation}
\begin{aligned}
c_{t+1:t+H}^{\prime} &= \text{IFFT}\left(\text{Filter}\left(\text{FFT}(c_{t+1:t+H})\right)\right) \\
\text{Filter}({\xi}) &= {\xi} \odot {\theta}_{c}
\end{aligned}
\end{equation}
where \(\text{FFT}(\cdot)\) denotes the Fast Fourier Transform along the temporal dimension, \(\text{IFFT}(\cdot)\) represents the Inverse Fast Fourier Transform, \(\theta_{c}\) corresponds to the learnable filter parameters for the concatenated periodic components, and \(\odot\) indicates element-wise multiplication (Hadamard product).

Both the periodic basis \(Q\) and the adaptive filters are jointly trained through gradient backpropagation to effectively extract the underlying low-frequency periodic patterns. 

Notably, the base period length \(W\) is determined by the inherent characteristics of the dataset. For time series with 1-hour intervals, \(W=24\) for daily cycles or \(W=168\) for weekly patterns. This architecture combines explicit periodic modeling with frequency-aware filtering, demonstrating superior performance in capturing dominant cyclical trends while maintaining computational efficiency through its parameter-sharing mechanism.

In conclusion, FECF procedure is as follow:

\noindent \hspace{1em}\textup{1) Extract Residual Component:} Remove the periodic component \(c_{t-L+1:t}\) from the input \(x_{t-L+1:t}\) obtaining the residual component:
\begin{equation}
r_{t-L+1:t} = x_{t-L+1:t} - c_{t-L+1:t}
\end{equation}
\noindent \hspace{1em}\textup{2) Residual Prediction via SFPL Module:} Feed the residual component into the SFPL module to predict future residuals:
\begin{equation}
r_{t+1:t+H} = \text{SFPL}(r_{t-L+1:t})
\end{equation}
\noindent \hspace{1em}\textup{3) Reconstruct Prediction:} Combine the predicted residual with the filtered periodic component: 
\begin{equation}
\overline{x}_{t+1:t+H} = r_{t+1:t+H} + c_{t+1:t+H}^{\prime}
\end{equation}

\subsection{Segmented Frequency-domain Pattern Learning}
The SFPL technique is specifically designed to enhance the energy contribution of mid to high frequency components, enabling the prediction of aperiodic components and short-term fluctuations. Within the FreqCycle framework, SFPL operates on residual components \(r_{t-L+1:t}\) obtained after removing low-frequency periodic elements from the input. SFPL procedure is as follow:

\noindent \hspace{1em}\textup{1)} The input \(r_{t-L+1:t}\) is divided into \(s\) sub-segments via a sliding window, with zero-padding applied to both ends of each sub-segment. These processed sub-segments are then stacked to obtain segmented time feature \(R=\{r_1,r_2,\ldots,r_{s}\}\in\mathbb{R}^{s\times L\times D}\).

\noindent \hspace{1em}\textup{2)} Then FFT is applied to \(R\) along the temporal dimension to obtain stacked frequency represention \(F\in \mathbb{C}^{s\times \lfloor{\frac{L}{2}+1}\rfloor\times D}\), after which frequency-domain feature extraction is performed on \(F\). To consolidate critical frequency-domain information, an adaptive weighted summation is conducted across the sub-segments in \(F\), ultimately yielding the refined frequency-domain data \(f\in\mathbb{C}^{\lfloor{\frac{L}{2}+1}\rfloor\times D}\) with enhanced features.
\begin{equation}
\begin{aligned}
\Theta_{F} &= \{ \theta_1, \ldots, \theta_{{\text{s}}} \} \\
\theta_1', \ldots, \theta_{{\text{s}}}' &= \text{softmax}(\theta_1, \ldots, \theta_{{\text{s}}}) \\
\Theta'_{F} &= \{ \theta_1', \ldots, \theta_{{\text{s}}}' \} \\
F' &= F \odot \Theta'_{F} \\
f &= \sum_{i=1}^{{\text{s}}} F'(i)
\end{aligned}
\label{eq:frequency_processing}
\end{equation}
where \(\Theta_{F}\) corresponds to the learnable parameters for the segmented frequency components. \(F'(i)\) referring to each sub-element of \(F\) along the segmented dimension.

\noindent \hspace{1em}\textup{3)} The refined data \(f\) undergoes iFFT and is then passed through FFN layer to produce the residual prediction output \(r_{t+1:t+H}\) of SFPL.

The segmentation operation is an important technique in SFPL. By dividing the signal into short sub-segments and performing FFT on each segment individually, it essentially constitutes a variant of the Short-Time Fourier Transform (STFT). Each sub-segment achieves higher frequency-domain locality by shortening the time-domain window \cite{STFT}, thereby enabling more precise localization of transient frequency components. The resulting frequency-domain representation \(f\), obtained through learnable filtering and adaptive weighting integration, becomes more discriminative with enhanced mid to high frequency energy. We provide a more detailed introduction to the theory of STFT and its strategies for enhancing mid to high frequency energy proportion in Appendix A.
\begin{table*}[htbp]
\centering
\setlength{\tabcolsep}{1.5pt} % 紧凑列间距
\renewcommand{\arraystretch}{1.2}
\footnotesize
\begin{tabular}{@{}l*{20}{c}@{}}
\toprule
Models&\multicolumn{2}{c}{FreqCycle} & \multicolumn{2}{c}{DLinear} & \multicolumn{2}{c}{CycleNet} & \multicolumn{2}{c}{Amplifier} & \multicolumn{2}{c}{FreTS} & 
\multicolumn{2}{c}{FilterNet} & \multicolumn{2}{c}{FITS} & \multicolumn{2}{c}{iTransformer} & \multicolumn{2}{c}{PatchTST} & \multicolumn{2}{c}{TimeMixer} \\
 \cmidrule(lr){2-3} \cmidrule(lr){4-5} \cmidrule(lr){6-7} \cmidrule(lr){8-9} \cmidrule(lr){10-11} 
\cmidrule(lr){12-13} \cmidrule(lr){14-15} \cmidrule(lr){16-17} \cmidrule(lr){18-19} \cmidrule(l){20-21}
 Metrics& MSE & MAE & MSE & MAE & MSE & MAE & MSE & MAE & MSE & MAE & MSE & MAE & MSE & MAE & MSE & MAE & MSE & MAE & MSE & MAE \\
\midrule
ETTm1 & \firstplace{0.372} & \firstplace{0.389} & 0.403 & 0.407 & 0.386 & 0.395 & \secondplace{0.381} & \secondplace{0.394} & 0.408 & 0.416 & 0.384 & 0.398 & 0.415 & 0.408 & 0.407 & 0.410 & 0.387 & 0.400 & 0.381 & 0.395 \\
\hline
ETTm2 & \firstplace{0.263} & \firstplace{0.311} & 0.350 & 0.401 & \secondplace{0.272} & \secondplace{0.315} & 0.276 & 0.323 & 0.321 & 0.368 & 0.276 & 0.322 & 0.286 & 0.328 & 0.288 & 0.332 & 0.281 & 0.326 & 0.275 & 0.323 \\
\hline                                       
ETTh1 & \firstplace{0.428} & \firstplace{0.427} & 0.456 & 0.452 & 0.432 & 0.427 & \secondplace{0.430} & \secondplace{0.428} & 0.475 & 0.463 & 0.440 & 0.432 & 0.451 & 0.440 & 0.454 & 0.448 & 0.469 & 0.455 & 0.447 & 0.440 \\
\hline
ETTh2 & 0.371 & 0.399 & 0.559 & 0.515 & 0.383 & 0.404 & \firstplace{0.359} & \firstplace{0.391} & 0.472 & 0.465 & 0.378 & 0.404 & 0.383 & 0.408 & 0.383 & 0.407 & 0.387 & 0.407 & \secondplace{0.364} & \secondplace{0.395} \\
\hline
Weather & \secondplace{0.243} & \firstplace{0.270} & 0.265 & 0.317 & 0.254 & 0.279 & 0.243 & \secondplace{0.271} & 0.250 & 0.270 & 0.245 & 0.272 & 0.249 & 0.276 & 0.258 & 0.278 & 0.259 & 0.281 & \firstplace{0.240} & 0.271 \\
\hline
ECL & \firstplace{0.168} & \firstplace{0.259} & 0.212 & 0.300 & \secondplace{0.170} & \secondplace{0.260} & 0.171 & 0.265 & 0.189 & 0.278 & 0.201 & 0.285 & 0.217 & 0.295 & 0.178 & 0.270 & 0.205 & 0.290 & 0.182 & 0.272 \\
\hline
Traffic & \secondplace{0.448} & \firstplace{0.261} & 0.625 & 0.383 & 0.485 & 0.313 & 0.482 & 0.315 & 0.618 & 0.390 & 0.521 & 0.340 & 0.627 & 0.376 & \firstplace{0.428} & \secondplace{0.282} & 0.481 & 0.304 & 0.484 & 0.297 \\

\bottomrule
\end{tabular}

\caption{Time series forecasting comparison. We set the lookback window size \(L\) as 96 and the prediction length as \(H \in \{96, 192, 336, 720\}\). The best results are highlighted in bold and the second best are underlined. Results are averaged from all prediction lengths. Full results for all datasets are listed in Table 2 of Appendix B.}
\label{table1}
\end{table*}
\subsection{MFreqCycle}
MFreqCycle is a multi-scale parallel framework that achieves high-precision forecasting by collaboratively extracting temporal features across different time scales. The features include periodic patterns (such as daily and weekly cycles) and non-periodic components. Current experiments have demonstrated the feasibility of incorporating weekly cycles on several datasets, while longer-period patterns like annual cycles remain unverified due to dataset limitations. The architecture comprises two key modules: a base cycle learning and frequency enhancement module, which focuses on capturing the smallest significant cycle (e.g., daily cycle) and its associated non-periodic features, and a weekly cycle learning and frequency enhancement module, which models macro-scale periodic patterns (e.g., weekly cycle). These multi-scale predictions are then adaptively integrated through a prediction fusion layer. To improve computational efficiency, the framework employs different input window lengths when extracting temporal features based on base-cycle and weekly-cycle durations.

To capture macro-scale periodic patterns, the system requires longer input sequences spanning, which significantly increases computational demands. However, for short-term forecasting, the model only needs to extract key trend information from data. To balance computational efficiency with model performance, the weekly cycle learning and frequency enhancement module incorporates three submodules: a pooling layer, a feature learning module, and a Linear layer. The distinctive feature extraction architecture based on pooling and Linear projection represents the key difference between the weekly cycle module and the base cycle module, while the feature learning submodule follows the same processing pipeline as FreqCycle. This design enables efficient extraction of essential weekly patterns without compromising predictive accuracy.

After processing through the base and weekly modules, the model generates multi-scale prediction outputs including the base period output \(\overline{x}_{t+1:t+H}^0\) and weekly period output \(\overline{x}_{t+1:t+H}^1\). Then the prediction fusion module performs adaptive integration of these multi-scale predictions to produce the final system output \(\overline{x}_{t+1:t+H}\).
\begin{equation}
\begin{aligned}
\theta_0', \theta_1' = &\text{softmax}(\theta_0, \theta_1)\\
 \overline{x}_{t+1:t+H} = \overline{x}_{t+1:t+H}^0& \odot \theta_0' +  \overline{x}_{t+1:t+H}^1 \odot \theta_1'
 \end{aligned}
\end{equation}
where \(\theta_0', \theta_1'\) corresponds to the learnable parameters for the prediction fusion module.

This fusion strategy ensures an optimal balance between different periodic features while maintaining computational efficiency and prediction accuracy. 

\section{Experiments}
\subsection{Experimental Setup}
\subsubsection{Datasets}
To validate the model's performance and generalization capability, we conduct experiments on seven datasets, including publicly benchmarks like ETT dataset series \cite{zhou2021informer}, Electricity (ECL), Weather and Traffic datasets \cite{autoformer}. These datasets span diverse domains with varying scales and sampling frequencies, providing a comprehensive testbed to evaluate the model's adaptability and scalability across different forecasting tasks. Detailed dataset information is summarized in Appendix B.
\subsubsection{Technical Implementation Details}
All experiments were implemented in PyTorch 2.6.0 and conducted on an NVIDIA RTX 4090 GPU with 24GB of memory.
\subsubsection{Baselines}
In this work, we benchmark our model against the representative and state-of-the-art approaches. We are evaluating and comparing with Linear and MLP-based models, including DLinear \cite{DLinear}, CycleNet \cite{cyclenet}, and TimeMixer \cite{timemixer}. We further evaluate against frequency learning methods, including FreTS \cite{yi2023frets}, FilterNet \cite{yi2024filternet}, FITS \cite{xufits}, and Amplifer \cite{fei2025amplifier}. To demonstrate the effectiveness of our model, we conduct comprehensive comparisons with Transformer-based models, including iTransformer \cite{iTransformer} and PatchTST \cite{PatchTST}.
\subsubsection{Evaluation Metrics}
Following established practices in prior work, we employ two primary metrics to assess forecasting performance: Mean Squared Error (MSE) and Mean Absolute Error (MAE).

\subsection{Main Results}
Table \ref{table1} presents the training results of the FreqCycle model and nine baselines on seven open-source datasets. Overall, FreqCycle secured first place in 10 out of 14 metrics and second place in 2 metrics. Compared to other models, our model demonstrates significant performance improvements, which we attribute to the FECF's learning of periodic patterns and the SFPL's enhancement of key frequencies.

To evaluate the effectiveness of the MFreqCycle model, we design Table \ref{table2} with lookback windows of 96 and a whole week length data 672(168), comparing the model against CycleNet and FITS. As shown in Table \ref{table2}, simply increasing the lookback window may lead to minor performance improvements or even slight degradation in some cases. The reason could be that the inherent limitations of the models prevent them from handling the complex temporal dependencies or noise interference introduced by longer lookback windows. However, the MFreqCycle model, specifically designed for longer lookback windows, achieves remarkable performance improvements on the ETT datasets, demonstrating its pioneering capability in learning both long-period patterns and their aperiodic components.
\begin{table}[tb]
\centering
\setlength{\tabcolsep}{4.5pt}
\small
\begin{tabular}{@{}lc|cc|cc|cc@{}}
\toprule
\multirow{2}{*}{Datasets} & \multirow{2}{*}{L} & \multicolumn{2}{c}{FreqCycle} & \multicolumn{2}{c}{CycleNet} & \multicolumn{2}{c}{FITS} \\
\cmidrule(lr){3-4} \cmidrule(lr){5-6} \cmidrule(lr){7-8}
 & & MSE & MAE & MSE & MAE & MSE & MAE \\
\midrule
ETTh2 & 96 & 0.371 & 0.399 & 0.383 & 0.404 & 0.383 & 0.408 \\
ETTm2 & 96 & 0.263 & 0.311 & 0.267 & 0.314 & 0.286 & 0.328 \\
\midrule
ETTh2 & 168 & \textbf{0.241} & \textbf{0.340} & 0.388 & 0.415 & 0.333 & 0.382 \\
ETTm2 & 672 & \textbf{0.167} & \textbf{0.279} & 0.262 & 0.325 & 0.250 & 0.312 \\
\bottomrule
\end{tabular}
\caption{Performance comparison with different lookback window lengths. The prediction length is \(H \in \{96, 192, 336, 720\}\). The best results are highlighted in bold. Results are averaged across all prediction lengths. Full results are available in Table 7 of Appendix D.}
\label{table2}
\end{table}

\subsection{Model Analysis}
\subsubsection{Ablation Study}
To validate the effectiveness of the FECF and SFPL modules in FreqCycle, we designed the ablation experiments shown in Table \ref{table3}. We compared the performance of: 1) the complete FreqCycle framework, 2) the variant without the FECF module, and 3) the variant replacing SFPL with MLP. The results show that both modules significantly contribute to improving prediction performance. Specifically, the FECF module achieves the most substantial improvements on Traffic and Weather datasets with prominent cycles, showing a 21.30\% MAE improvement on the Traffic dataset. The SFPL module exhibits notable performance gains on ETTh datasets, which exhibit stronger non-stationary characteristics and richer transient frequency components, achieving a 6.91\% MSE improvement. SFPL and FECF demonstrate complementary performance characteristics, and their synergistic integration enables the FreqCycle model to achieve superior predictive performance.
\begin{table}[tb]
\centering
\setlength{\tabcolsep}{0.01pt} % Compact column spacing
\small
\begin{tabular}{lcccccccc}
\toprule
Datasets & \multicolumn{2}{c}{ETTh1} & \multicolumn{2}{c}{ETTh2} & \multicolumn{2}{c}{Traffic} & \multicolumn{2}{c}{Weather} \\
\cmidrule(lr){2-3} \cmidrule(lr){4-5} \cmidrule(lr){6-7} \cmidrule(lr){8-9}
Metrics & MSE & MAE & MSE & MAE & MSE & MAE & MSE & MAE \\
\midrule
FreqCycle & 0.369 & 0.390 & 0.282 & 0.336 & 0.438 & 0.262 & 0.159 & 0.203 \\
No FECF & 0.375 & 0.397 & 0.292 & 0.344 & 0.501 & 0.318 & 0.179 & 0.218 \\
Boost & \textbf{1.65\%} & \textbf{1.82\%} & \textbf{3.26\%} & \textbf{2.56\%} & \textbf{14.54\%} & \textbf{21.30\%} & \textbf{12.59\%} & \textbf{7.29\%} \\
2MLP & 0.379 & 0.401 & 0.302 & 0.348 & 0.444 & 0.262 & 0.161 & 0.206 \\
Boost & \textbf{2.66\%} & \textbf{2.93\%} & \textbf{6.91\%} & \textbf{3.85\%} & \textbf{1.44\%} & \textbf{0.04\%} & \textbf{1.51\%} & \textbf{1.28\%} \\
\bottomrule
\end{tabular}
\caption{Ablation study of FreqCycle module, using parameters L=96 and H=96. 2MLP means the variant replacing SFPL with MLP. }
\label{table3}
\end{table}

\subsubsection{Effectiveness of SFPL}
In recent years, researchers have introduced signal processing techniques, like low-pass filters (LPFs), into time series forecasting and developed learnable filter schemes, aiming to extract more critical frequency components. Unlike conventional LPFs with rigid cutoff frequencies or purely gradient-based learnable filters, our approach draws inspiration from STFT to enhance the extraction of crucial mid to high frequency information.

We conduct a systematic comparison between our SFPL module and existing frequency processing methods within the FreqCycle framework. We evaluate: 1) replacing SFPL's frequency processing with PaiFilter \cite{yi2024filternet}, and 2) substituting SFPL with the LPF scheme from FITS \cite{xufits}. As shown in Table \ref{table4}, these results indicate that existing frequency-domain filtering approaches fail to adequately capture critical frequency components, whereas our SFPL module effectively extracts frequency features.
\begin{table}[tb]
\centering
\setlength{\tabcolsep}{1.7pt} % Compact column spacing
\small
\begin{tabular}{lcccccc}
\toprule
Datasets & \multicolumn{2}{c}{ETTh2} & \multicolumn{2}{c}{ECL} & \multicolumn{2}{c}{Traffic} \\
\cmidrule(lr){2-3} \cmidrule(lr){4-5} \cmidrule(lr){6-7}
Metrics & MSE & MAE & MSE & MAE & MSE & MAE \\
\midrule
FreqCycle & 0.282 & 0.336 & 0.139 & 0.233 & 0.438 & 0.262 \\
SFPL2LPF & 0.294 & 0.345 & 0.143 & 0.240 & 0.449 & 0.268 \\
Boost & \textbf{4.00\%} & \textbf{2.68\%} & \textbf{3.39\%} & \textbf{3.09\%} & \textbf{2.63\%} & \textbf{2.29\%} \\
SFPL2PaiFilter & 0.293 & 0.344 & 0.143 & 0.239 & 0.445 & 0.262 \\
Boost & \textbf{3.68\%} & \textbf{2.59\%} & \textbf{2.96\%} & \textbf{2.58\%} & \textbf{1.81\%} & \textbf{0.08\%} \\
\bottomrule
\end{tabular}
\caption{Comparison of frequency processing methods, using parameters L=96 and H=96.}
\label{table4}
\end{table}

\subsubsection{Effectiveness of FECF}
Our FECF technique fundamentally operates as a Seasonal-Trend Decomposition (STD) method. To validate its effectiveness, we conducted experiments comparing it against two established STD approaches: the moving average decomposition (MOV) used in DLinear \cite{DLinear} and the local decomposition (LD) \cite{yu2024revitalizing}. As demonstrated in Table \ref{table5}, when replacing FECF with either MOV+Linear (DFreq) or LD+Linear (LDFreq) variants in our FreqCycle framework, the forecasting performance consistently degraded across all seven benchmark datasets. These empirical results clearly establish that our FECF-based decomposition achieves superior temporal pattern extraction compared to conventional STD methods.
\begin{table}[htbp]
\centering
\setlength{\tabcolsep}{4pt}
\footnotesize
\begin{tabular}{lcccccc}
\toprule
Datasets & \multicolumn{2}{c}{FreqCycle} & \multicolumn{2}{c}{DFreq} & \multicolumn{2}{c}{LDFreq} \\
\cmidrule(lr){2-3} \cmidrule(lr){4-5} \cmidrule(lr){6-7}
 Metrics& MSE & MAE & MSE & MAE & MSE & MAE \\
\midrule
ETTm2 & \textbf{0.263} & \textbf{0.311} & 0.312 & 0.365 & 0.329 & 0.368 \\
Weather & \textbf{0.243} & \textbf{0.270} & 0.259 & 0.309 & 0.248 & 0.291 \\
ECL & \textbf{0.168} & \textbf{0.259} & 0.204 & 0.293 & 0.201 & 0.288 \\
Traffic & \textbf{0.448} & \textbf{0.261} & 0.529 & 0.316 & 0.527 & 0.307 \\
\bottomrule
\end{tabular}
\caption{Performance comparison of different STD methods. The configuration look-back window length of 96 and The reported results are averaged across all prediction horizons of \(H \in \{96, 192, 336, 720\}\), with full results available in Table 5 of Appendix D. }
\label{table5}
\end{table}

\begin{figure}[tb]
\centering
\includegraphics[width=0.99\columnwidth]{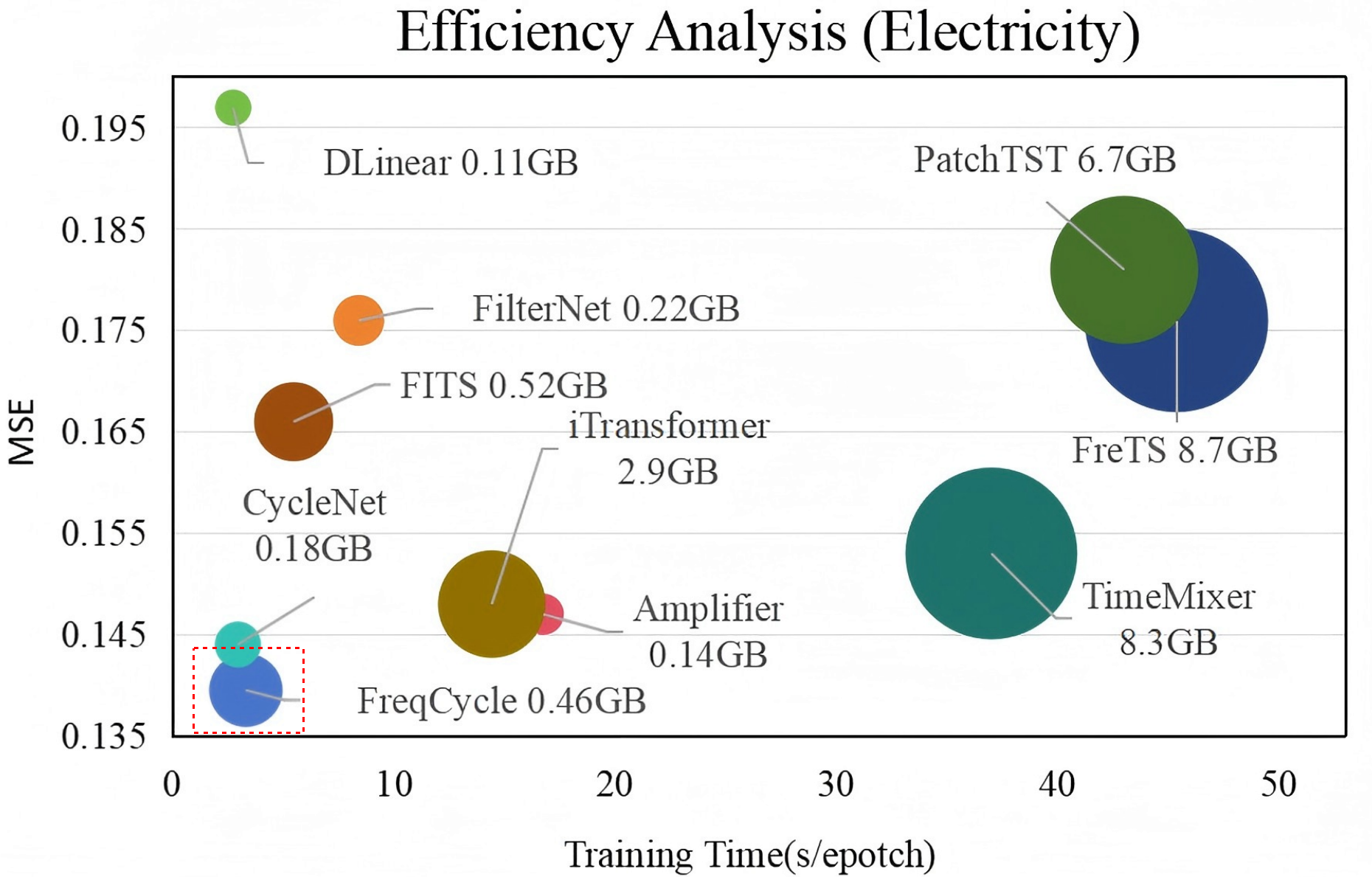}
\caption{Model efficiency comparison in terms of MSE, Peak memory consumption and Training speed. Using an input length \(L=96\) and prediction horizon \(H=96\) on the Electricity dataset.}
\label{figeffency}
\end{figure}
\subsubsection{Efficiency Analysis}
The theoretical complexity of FreqCycle is \(O(L \log L)\). We conducted a comprehensive comparison with representative models across three dimensions, Forecasting performance, Peak memory consumption and Training speed.
Figure \ref{figeffency} demonstrates that FreqCycle achieves the optimal balance between model performance and computational efficiency.
This efficiency advantage stems from our time and frequency processing paradigm, which avoids the quadratic complexity of attention mechanisms while preserving temporal pattern learning capabilities.

\subsection{Visualization}
\subsubsection{Visualization of SFPL's Key Frequency Enhancement}
\begin{figure}[tb]
\centering
\includegraphics[width=1\columnwidth]{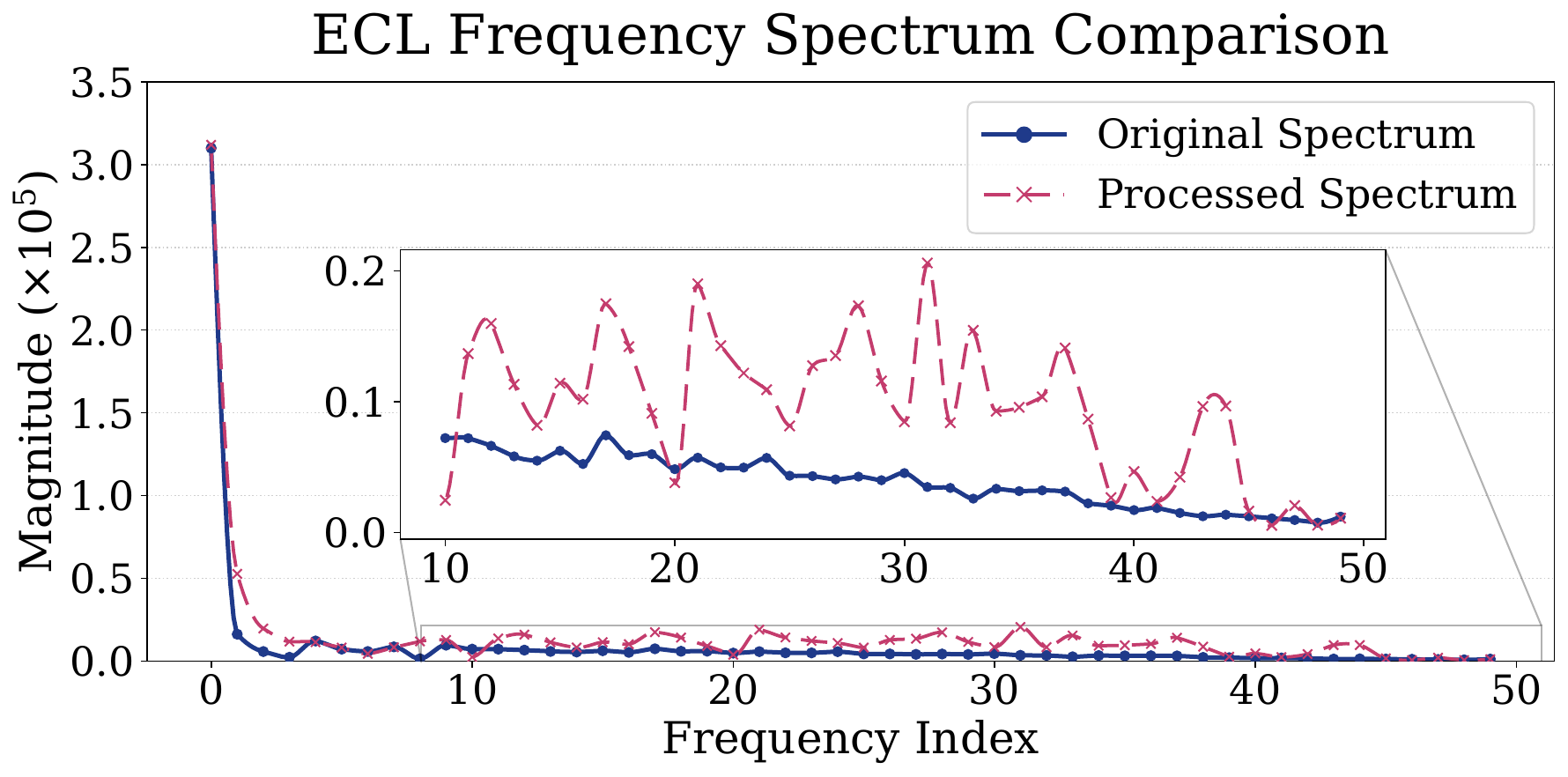}
\caption{Frequency spectrum of Electricity dataset before and
after SFPL enhancement. The other datasets' results are available in Appendix E.}
\label{figfrequency}
\end{figure}

After training the model on benchmark datasets, we analyzed the frequency amplitude spectra before and after SFPL processing. As demonstrated in Figure \ref{figfrequency}, SFPL effectively Amplifies the energy proportion of mid to high frequency bands as well as Preserves the structural integrity of dominant low-frequency components.
\subsubsection{Visualization of Periodic Patterns Extracted by FECF}
\begin{figure}[tb]
\centering
\includegraphics[width=1\columnwidth]{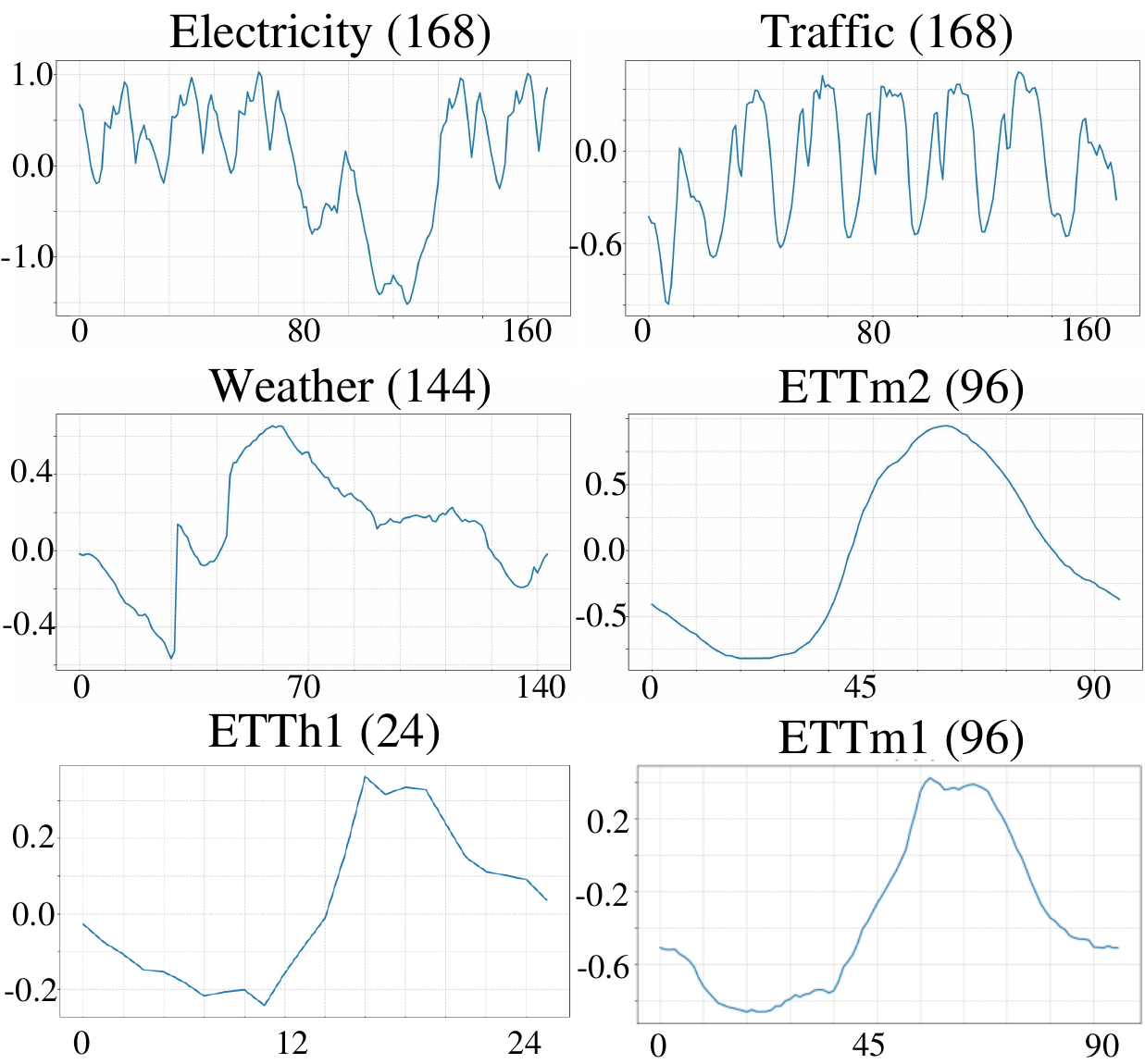} 
\caption{Learned periodic patterns by conducting hyperparameter \(W\) tuning on each dataset.}
\label{fig3}
\end{figure}

Figure \ref{fig3} shows the learned periodic patterns across multiple datasets after hyperparameter W optimization. Our analysis reveals that for the Electricity and Traffic datasets, FreqCycle successfully captures weekly periodicity patterns through hyperparameter tuning. For the remaining datasets, the optimized FreqCycle configuration learns daily cycle patterns.
\subsubsection{Visualization of Forecasting Results}
The prediction results on benchmark datasets are available in Appendix E.
\section{Conclusion}
This paper presents FreqCycle, a frequency aware framework for TSF that addresses the under explored challenge of modeling mid to high frequency components. FreqCycle integrates two key innovations, FECF for periodic pattern extraction, and SFPL to amplify mid to high frequency energy via learnable filters and adaptive weighting. For coupled multi-scale periodicity, we further propose MFreqCycle, which achieves remarkable performance improvements on several datasets while maintaining efficiency. Comprehensive experiments across seven real world datasets demonstrate both the superior accuracy and computational efficiency of our proposed approach. The FreqCycle framework establishes a new paradigm for time series forecasting that achieves state-of-the-art performance while maintaining lightweight architecture.
 
\section*{Acknowledgments}
We would like to express our sincere gratitude to our advisor, Associate Professor Xing He, for his continuous guidance and invaluable support throughout this research. 
We also thank all members of the Intelligent Information Processing and Control Laboratory (IIC) at Shanghai Jiao Tong University for their insightful discussions and kind assistance during the development of this work.

\bibliography{aaai2026}

\clearpage
%File: formatting-instructions-latex-2026.tex
%release 2026.0

%File: formatting-instructions-latex-2026.tex
%release 2026.0

\urlstyle{rm} % DO NOT CHANGE THIS
\def\UrlFont{\rm}  % DO NOT CHANGE THIS

\frenchspacing  % DO NOT CHANGE THIS
\setlength{\pdfpagewidth}{8.5in}  % DO NOT CHANGE THIS
\setlength{\pdfpageheight}{11in}  % DO NOT CHANGE THIS
%
% These are recommended to typeset algorithms but not required. See the subsubsection on algorithms. Remove them if you don't have algorithms in your paper.

%
% These are are recommended to typeset listings but not required. See the subsubsection on listing. Remove this block if you don't have listings in your paper.

\lstset{%
	basicstyle={\footnotesize\ttfamily},% footnotesize acceptable for monospace
	numbers=left,numberstyle=\footnotesize,xleftmargin=2em,% show line numbers, remove this entire line if you don't want the numbers.
	aboveskip=0pt,belowskip=0pt,%
	showstringspaces=false,tabsize=2,breaklines=true}
\floatstyle{ruled}
\newfloat{listing}{tb}{lst}{}
\floatname{listing}{Listing}
%
% Keep the \pdfinfo as shown here. There's no need
% for you to add the /Title and /Author tags.
\pdfinfo{
/TemplateVersion (2026.1)
}

\setcounter{secnumdepth}{0} %May be changed to 1 or 2 if section numbers 

% REMOVE THIS: bibentry
% This is only needed to show inline citations in the guidelines document. You should not need it and can safely delete it.

% END REMOVE bibentry

\section{Appendix}
\section{A. Introduction of Short-Time Fourier Transform}  
The short-time Fourier transform (STFT) is a widely used time-frequency representation for signals such as speech and images.  

For a discrete-time signal $x(n)$, the STFT is defined as  
\begin{equation}  
X_w(nL, \omega) = \sum_{m} x(m)w(nL - m)e^{-j\omega m}  
\label{eq:stft}  
\end{equation}  
where the subscript $w$ in $X_w(nL, \omega)$ denotes the analysis window $w(n)$. The parameter $L$ is an integer which denotes the separation in time between adjacent short-time sections. This parameter is independent of time and is selected so as to ensure a degree of time overlap between adjacent short-time sections.  

For a fixed value of $n$, $X_w(nL, \omega)$ represents the Fourier transform with respect to $m$ of the short-time section $f_n(m) = x(m)w(nL - m)$. The sliding window interpretation views $X_w(nL, \omega)$ as being generated by shifting the time-reversed analysis window across the signal. After each shift of $L$ samples, the window is multiplied with the signal and the Fourier transform is applied to the product.  

There are other interpretations of the STFT, including a well-known filter bank interpretation. For the purposes of our paper, we find the sliding window interpretation to be the most appropriate.

\subsubsection{Window Function Characteristics}
The choice of window function critically affects the STFT's spectral resolution and temporal localization. Common window functions include:

\begin{itemize}
    \item \textbf{Rectangular Window}:
    \begin{equation}
    w_{\text{rect}}(n) = 1 \quad \text{for} \quad 0 \leq n \leq N-1
    \end{equation}
    
    \item \textbf{Hann Window}:
    \begin{equation}
    w_{\text{hann}}(n) = 0.5\left(1 - \cos\left(\frac{2\pi n}{N-1}\right)\right)
    \end{equation}
    
    \item \textbf{Hamming Window}:
    \begin{equation}
    w_{\text{hamming}}(n) = 0.54 - 0.46\cos\left(\frac{2\pi n}{N-1}\right)
    \end{equation}
\end{itemize}

The sliding window perspective proves particularly advantageous for our analysis of time-varying spectral characteristics in time series. Window selection involves trade-offs between main lobe width and side lobe attenuation, with Hann windows offering good frequency resolution and Hamming windows providing superior side lobe suppression.

\subsection{STFT-Based Mid to High Frequency Enhancement via Window Adjustment}
\subsubsection{Time-Frequency Resolution Trade-off}
The fundamental limitation of STFT is governed by the Heisenberg uncertainty principle, which restricts the joint time-frequency resolution. This trade-off can be expressed as:
\begin{equation}
\Delta t \cdot \Delta \omega \geq \frac{1}{4\pi}
\end{equation}
where $\Delta t$ and $\Delta \omega$ represent time and frequency resolution respectively.

\subsubsection{Window Length Selection Strategy}
For mid to high frequency component analysis, shorter windows are preferred as they:
\begin{itemize}

\item Naturally amplify mid to high frequency energy through spectral redistribution.
\item Provide better temporal localization for transients.
\end{itemize}

\subsubsection{Implementation Considerations}
The practical implementation involves:
\begin{enumerate}
\item Window Function Choice
\item Length Adjustment: Longer windows improve frequency resolution while shorter windows enhance temporal localization.
\end{enumerate}

\section{B. Supplementary Details of the Experiments}  
\subsubsection{Experiment Settings}
We utilized widely used benchmark datasets for LTSF tasks, including the ETT series, Electricity, Traffic, and Weather. We split the ETTs dataset and traffic dataset into training, validation, and test sets with a ratio of 6:2:2, while the other datasets were split in a ratio of 7:1:2.
We implementing all experiments in PyTorch 2.6.0 on an NVIDIA RTX 4090 GPU with 24GB of memory. The models are trained for 30 epochs using Adam optimizer with learning rates in \( \{0.005, 0.01, 0.025, 0.03\}\), batch sizes in \(\{64, 128, 256\}\), random seeds in \(\{1,2,2024,2025,2026\}\) and hidden dimensions in \(\{128, 512\}\), employing ReLU activation. The code is included as supplementary material and will be made publicly available.
\subsubsection{Datasets}
\begin{itemize}
    \item \textbf{ETT} dataset contains two sub-datasets:
    \begin{itemize}
        \item ETTh: Collected at 1-hour intervals
        \item ETTm: Collected at 15-minute intervals
    \end{itemize}
    Both were recorded from electricity transformers between July 2016 and July 2018.
    
    \item \textbf{Electricity} records hourly consumption of 321 clients from 2012 to 2014.
        
    \item \textbf{Traffic} provides hourly measurements from 862 sensors on San Francisco Bay Area freeways since January 2015.
    
    \item \textbf{Weather} includes 21 meteorological indicators recorded every 10 minutes during 2020 at the Max Planck Biogeochemistry Institute weather station.
\end{itemize}

The detailed dataset statistics are summarized in Table~\ref{tab:dataset_stats}.

\begin{table}[t]
\centering

\begin{tabular}{lcccc}
\toprule
Datasets & Frequency & Duration & Features & Samples \\
\midrule
ETTh & 1 hour & 2016-2018 & 7 & 17420 \\
ETTm & 15 min & 2016-2018 & 7 & 69680 \\
Electricity & 1 hour & 2012-2014 & 321 & 26,304 \\
Traffic & 1 hour & 2015-2018 & 862 & 17,544 \\
Weather & 10 min & 2020 & 21 & 52,696 \\
\bottomrule
\end{tabular}
\caption{Dataset Specifications}
\label{tab:dataset_stats}
\end{table}
\subsubsection{Full Results}
The full multivariate time series forecasting results of FreqCycle are presented in Table \ref{table1}, along with extensive evaluations of competitive counterparts. 
\begin{table*}[htbp]
\centering
% Adjust column spacing
\setlength{\tabcolsep}{1pt} % Compact column spacing
\renewcommand{\arraystretch}{0.8} % Compact row spacing
\small
\setlength{\tabcolsep}{1.3pt}
\renewcommand{\arraystretch}{1.1}

\begin{tabular}{@{}l l *{10}{|cc} @{}}
\toprule
\multirow{2}{*}{\rotatebox[origin=c]{90}{Dataset}} & \multirow{2}{*}{H} & 
\multicolumn{2}{c}{FreqCycle} & \multicolumn{2}{c}{DLinear} & \multicolumn{2}{c}{CycleNet} & \multicolumn{2}{c}{Amplifier} & 
\multicolumn{2}{c}{FreTS} & \multicolumn{2}{c}{FilterNet} & \multicolumn{2}{c}{FITS} & \multicolumn{2}{c}{iTransformer} & 
\multicolumn{2}{c}{PatchTST} & \multicolumn{2}{c@{}}{TimeMixer} \\
\cmidrule(lr){3-4} \cmidrule(lr){5-6} \cmidrule(lr){7-8} \cmidrule(lr){9-10} \cmidrule(lr){11-12} 
\cmidrule(lr){13-14} \cmidrule(lr){15-16} \cmidrule(lr){17-18} \cmidrule(lr){19-20} \cmidrule(lr){21-22}
 & & \multicolumn{1}{c}{MSE} & \multicolumn{1}{c}{MAE} & \multicolumn{1}{c}{MSE} & \multicolumn{1}{c}{MAE} & \multicolumn{1}{c}{MSE} & \multicolumn{1}{c}{MAE} & \multicolumn{1}{c}{MSE} & \multicolumn{1}{c}{MAE} & 
 \multicolumn{1}{c}{MSE} & \multicolumn{1}{c}{MAE} & \multicolumn{1}{c}{MSE} & \multicolumn{1}{c}{MAE} & \multicolumn{1}{c}{MSE} & \multicolumn{1}{c}{MAE} & \multicolumn{1}{c}{MSE} & \multicolumn{1}{c}{MAE} & 
 \multicolumn{1}{c}{MSE} & \multicolumn{1}{c}{MAE} & \multicolumn{1}{c}{MSE} & \multicolumn{1}{c@{}}{MAE} \\
\midrule

\multirow{5}{*}{\rotatebox[origin=c]{90}{ETTh1}} 
& 96 & \firstplace{0.369} & \firstplace{0.390} & 0.386 & 0.400 & 0.378 & 0.391 & \secondplace{0.371} & \secondplace{0.392} & 0.395 & 0.407 & 0.375 & 0.394 & 0.386 & 0.396 & 0.386 & 0.405 & 0.414 & 0.419 & 0.375 & 0.400 \\
& 192 & \firstplace{0.417} & \firstplace{0.418} & 0.437 & 0.432 & 0.426 & 0.419 & \secondplace{0.426} & \secondplace{0.422} & 0.448 & 0.440 & 0.436 & 0.422 & 0.436 & 0.423 & 0.441 & 0.436 & 0.460 & 0.445 & 0.429 & 0.421 \\
& 336 & \secondplace{0.459} & \secondplace{0.439} & 0.481 & 0.459 & 0.464 & 0.439 & \firstplace{0.448} & \firstplace{0.434} & 0.499 & 0.472 & 0.476 & 0.443 & 0.478 & 0.444 & 0.487 & 0.458 & 0.501 & 0.466 & 0.484 & 0.458 \\
& 720 & \firstplace{0.456} & \firstplace{0.454} & 0.519 & 0.516 & 0.461 & 0.460 & \secondplace{0.476} & \secondplace{0.464} & 0.558 & 0.532 & 0.474 & 0.469 & 0.502 & 0.495 & 0.503 & 0.491 & 0.500 & 0.488 & 0.498 & 0.482 \\
\cmidrule(lr){2-22}
& AVG & \firstplace{0.425} & \firstplace{0.425} & 0.456 & 0.452 & 0.432 & 0.427 & \secondplace{0.430} & \secondplace{0.428} & 0.475 & 0.463 & 0.440 & 0.432 & 0.451 & 0.440 & 0.454 & 0.448 & 0.469 & 0.455 & 0.447 & 0.440 \\
\midrule

\multirow{5}{*}{\rotatebox[origin=c]{90}{ETTh2}}
& 96 & \secondplace{0.282} & \firstplace{0.336} & 0.333 & 0.387 & 0.285 & 0.335 & \firstplace{0.279} & \secondplace{0.337} & 0.309 & 0.364 & 0.292 & 0.343 & 0.295 & 0.350 & 0.297 & 0.349 & 0.302 & 0.348 & 0.289 & 0.341 \\
& 192 & \secondplace{0.361} & \firstplace{0.388} & 0.477 & 0.476 & 0.373 & 0.391 & \firstplace{0.359} & \secondplace{0.389} & 0.395 & 0.425 & 0.369 & 0.395 & 0.381 & 0.396 & 0.380 & 0.400 & 0.388 & 0.400 & 0.372 & 0.392 \\
& 336 & 0.415 & 0.428 & 0.594 & 0.541 & 0.421 & 0.433 & \firstplace{0.377} & \firstplace{0.406} & 0.462 & 0.467 & 0.420 & 0.432 & 0.426 & 0.438 & 0.428 & 0.432 & 0.426 & 0.433 & \secondplace{0.386} & \secondplace{0.414} \\
& 720 & 0.425 & 0.443 & 0.831 & 0.657 & 0.453 & 0.458 & \secondplace{0.420} & \firstplace{0.432} & 0.721 & 0.604 & 0.430 & 0.446 & 0.431 & 0.446 & 0.427 & 0.445 & 0.431 & 0.446 & \firstplace{0.412} & \secondplace{0.434} \\
\cmidrule(lr){2-22}
& AVG & 0.371 & 0.399 & 0.559 & 0.515 & 0.383 & 0.404 & \firstplace{0.359} & \firstplace{0.391} & 0.472 & 0.465 & 0.378 & 0.404 & 0.383 & 0.408 & 0.383 & 0.407 & 0.387 & 0.407 & \secondplace{0.365} & \secondplace{0.395} \\
\midrule

\multirow{5}{*}{\rotatebox[origin=c]{90}{ETTm1}}
& 96 & \firstplace{0.310} & \firstplace{0.353} & 0.345 & 0.372 & 0.325 & 0.363 & \secondplace{0.316} & \secondplace{0.355} & 0.335 & 0.372 & 0.318 & 0.358 & 0.355 & 0.375 & 0.334 & 0.368 & 0.329 & 0.367 & 0.320 & 0.357 \\
& 192 & \firstplace{0.351} & \firstplace{0.374} & 0.380 & 0.389 & 0.366 & 0.382 & \secondplace{0.361} & \secondplace{0.381} & 0.388 & 0.401 & 0.364 & 0.383 & 0.392 & 0.393 & 0.377 & 0.391 & 0.367 & 0.385 & \secondplace{0.361} & \secondplace{0.381} \\
& 336 & \firstplace{0.383} & \firstplace{0.396} & 0.413 & 0.413 & 0.396 & 0.401 & 0.393 & \secondplace{0.402} & 0.421 & 0.426 & 0.396 & 0.406 & 0.424 & 0.414 & 0.426 & 0.420 & 0.399 & 0.410 & \secondplace{0.390} & 0.404 \\
& 720 & \firstplace{0.446} & \firstplace{0.434} & 0.474 & 0.453 & 0.457 & 0.433 & 0.455 & 0.440 & 0.486 & 0.465 & 0.456 & 0.444 & 0.487 & 0.449 & 0.491 & 0.459 & 0.454 & \secondplace{0.439} & \secondplace{0.454} & 0.441 \\
\cmidrule(lr){2-22}
& AVG & \firstplace{0.372} & \firstplace{0.389} & 0.403 & 0.407 & 0.386 & 0.395 & \secondplace{0.381} & \secondplace{0.395} & 0.408 & 0.416 & 0.384 & 0.398 & 0.415 & 0.408 & 0.407 & 0.410 & 0.387 & 0.400 & \secondplace{0.381} & \secondplace{0.396} \\
\midrule

\multirow{5}{*}{\rotatebox[origin=c]{90}{ETTm2}}
& 96 & \firstplace{0.162} & \firstplace{0.244} & 0.193 & 0.292 & \secondplace{0.166} & \secondplace{0.248} & 0.176 & 0.258 & 0.189 & 0.277 & 0.174 & 0.257 & 0.183 & 0.266 & 0.180 & 0.264 & 0.175 & 0.259 & 0.175 & 0.258 \\
& 192 & \firstplace{0.226} & \firstplace{0.286} & 0.284 & 0.362 & \secondplace{0.233} & \secondplace{0.291} & 0.239 & 0.300 & 0.258 & 0.326 & 0.240 & 0.300 & 0.247 & 0.305 & 0.250 & 0.309 & 0.241 & 0.302 & 0.237 & 0.299 \\
& 336 & \firstplace{0.283} & \firstplace{0.326} & 0.369 & 0.427 & \secondplace{0.293} & \secondplace{0.330} & 0.297 & 0.338 & 0.343 & 0.390 & 0.297 & 0.339 & 0.307 & 0.342 & 0.311 & 0.348 & 0.305 & 0.343 & 0.298 & 0.340 \\
& 720 & \firstplace{0.383} & \firstplace{0.387} & 0.554 & 0.522 & 0.395 & \secondplace{0.389} & \secondplace{0.393} & 0.396 & 0.495 & 0.480 & 0.392 & 0.393 & 0.407 & 0.399 & 0.412 & 0.407 & 0.402 & 0.400 & 0.391 & 0.396 \\
\cmidrule(lr){2-22}
& AVG & \firstplace{0.263} & \firstplace{0.311} & 0.350 & 0.401 & \secondplace{0.272} & \secondplace{0.315} & 0.276 & 0.323 & 0.321 & 0.368 & 0.276 & 0.322 & 0.286 & 0.328 & 0.288 & 0.332 & 0.281 & 0.326 & 0.275 & 0.323 \\
\midrule

\multirow{5}{*}{\rotatebox[origin=c]{90}{Weather}}
& 96 & \secondplace{0.159} & \firstplace{0.203} & 0.196 & 0.255 & 0.170 & 0.216 & \firstplace{0.156} & \secondplace{0.204} & 0.174 & 0.208 & 0.164 & 0.207 & 0.166 & 0.213 & 0.174 & 0.214 & 0.177 & 0.218 & 0.163 & 0.209 \\
& 192 & \firstplace{0.208} & \firstplace{0.248} & 0.237 & 0.296 & 0.222 & 0.259 & \secondplace{0.209} & \secondplace{0.249} & 0.219 & 0.250 & 0.210 & 0.250 & 0.213 & 0.254 & 0.221 & 0.254 & 0.225 & 0.259 & \firstplace{0.208} & 0.250 \\
& 336 & \secondplace{0.264} & \secondplace{0.289} & 0.283 & 0.335 & 0.275 & 0.296 & 0.264 & 0.290 & 0.273 & 0.290 & 0.265 & 0.290 & 0.269 & 0.294 & 0.278 & 0.296 & 0.278 & 0.297 & \firstplace{0.251} & \firstplace{0.287} \\
& 720 & 0.343 & 0.342 & 0.345 & 0.381 & 0.349 & 0.345 & 0.343 & 0.342 & \firstplace{0.334} & \firstplace{0.332} & 0.342 & \secondplace{0.340} & 0.346 & 0.343 & 0.358 & 0.347 & 0.354 & 0.348 & \secondplace{0.339} & 0.341 \\
\cmidrule(lr){2-22}
& AVG & \secondplace{0.243} & \firstplace{0.270} & 0.265 & 0.317 & 0.254 & 0.279 & \secondplace{0.243} & \secondplace{0.271} & 0.250 & \firstplace{0.270} & 0.245 & 0.272 & 0.249 & 0.276 & 0.258 & 0.278 & 0.259 & 0.281 & \firstplace{0.240} & 0.272 \\
\midrule

\multirow{5}{*}{\rotatebox[origin=c]{90}{ECL}}
& 96 & \firstplace{0.140} & \firstplace{0.234} & 0.197 & 0.282 & \secondplace{0.141} & \secondplace{0.234} & 0.147 & 0.243 & 0.176 & 0.258 & 0.176 & 0.264 & 0.200 & 0.278 & 0.148 & 0.248 & 0.181 & 0.270 & 0.153 & 0.247 \\
& 192 & \firstplace{0.154} & \firstplace{0.246} & 0.196 & 0.285 & \secondplace{0.155} & \secondplace{0.247} & 0.157 & 0.251 & 0.175 & 0.262 & 0.185 & 0.270 & 0.200 & 0.280 & 0.162 & 0.253 & 0.188 & 0.274 & 0.166 & 0.256 \\
& 336 & \firstplace{0.170} & \firstplace{0.264} & 0.209 & 0.301 & \secondplace{0.172} & \secondplace{0.264} & 0.174 & 0.269 & 0.185 & 0.278 & 0.202 & 0.286 & 0.214 & 0.295 & 0.178 & 0.269 & 0.204 & 0.293 & 0.185 & 0.277 \\
& 720 & \secondplace{0.210} & \secondplace{0.297} & 0.245 & 0.333 & 0.210 & 0.296 & \firstplace{0.206} & \firstplace{0.296} & 0.220 & 0.315 & 0.242 & 0.319 & 0.255 & 0.327 & 0.225 & 0.317 & 0.246 & 0.324 & 0.225 & 0.310 \\
\cmidrule(lr){2-22}
& AVG & \firstplace{0.168} & \firstplace{0.260} & 0.212 & 0.300 & \secondplace{0.170} & \secondplace{0.260} & 0.171 & 0.265 & 0.189 & 0.278 & 0.201 & 0.285 & 0.217 & 0.295 & 0.178 & 0.272 & 0.205 & 0.290 & 0.182 & 0.273 \\
\midrule

\multirow{5}{*}{\rotatebox[origin=c]{90}{Traffic}}
& 96 & \secondplace{0.438} & \firstplace{0.262} & 0.650 & 0.396 & 0.480 & 0.314 & 0.455 & 0.298 & 0.593 & 0.378 & 0.506 & 0.336 & 0.651 & 0.391 & \firstplace{0.395} & \secondplace{0.268} & 0.462 & 0.295 & 0.462 & 0.285 \\
& 192 & \secondplace{0.434} & \firstplace{0.258} & 0.598 & 0.370 & 0.482 & 0.313 & 0.470 & 0.316 & 0.595 & 0.377 & 0.508 & 0.333 & 0.602 & 0.363 & \firstplace{0.417} & \secondplace{0.276} & 0.466 & 0.296 & 0.473 & 0.296 \\
& 336 & \secondplace{0.446} & \firstplace{0.258} & 0.605 & 0.373 & 0.476 & 0.303 & 0.479 & 0.316 & 0.609 & 0.385 & 0.518 & 0.335 & 0.609 & 0.366 & \firstplace{0.433} & \secondplace{0.283} & 0.482 & 0.304 & 0.498 & 0.296 \\
& 720 & \secondplace{0.476} & \firstplace{0.265} & 0.645 & 0.394 & 0.503 & 0.320 & 0.523 & 0.328 & 0.673 & 0.418 & 0.553 & 0.354 & 0.647 & 0.385 & \firstplace{0.467} & \secondplace{0.302} & 0.514 & 0.322 & 0.506 & 0.313 \\
\cmidrule(lr){2-22}
& AVG & \secondplace{0.448} & \firstplace{0.261} & 0.625 & 0.383 & 0.485 & 0.313 & 0.482 & 0.315 & 0.618 & 0.390 & 0.521 & 0.340 & 0.627 & 0.376 & \firstplace{0.428} & \secondplace{0.282} & 0.481 & 0.304 & 0.485 & 0.298 \\
\bottomrule

\bottomrule
\end{tabular}
\caption{Time series forecasting comparison. We set the lookback window size \(L\) as 96 and the prediction length as \(H \in \{96, 192, 336, 720\}\). The best results are in red and the second best are blue.}
\label{table1}
\end{table*}

\section{C. Window Strategy} 
As mentioned in Appendix A regarding the windowing strategy, this chapter conducts comparative experiments on FreqCycle in benchmarks from two perspectives: window type and window size.
\subsubsection{Window Type}
We choose Rectangular Window, Hann Window, and Hamming Window for SFPL's data segment, and the result can be seen in Table \ref{table_windowtype}. We uniformly set the window length to 6 and the stride to 6. Based on the experimental results, the rectangular window (rect) achieved optimal performance in over 85\% of test cases, primarily due to its ability to fully preserve the time-domain signal energy. The Hamming window (hamming) ranked second in performance due to its trade-off between time and frequency characteristics, exhibiting an average error 0.3\%-0.5\% higher than the rectangular window, while the Hann window (hann) performed the worst.

This suggests that in time-series forecasting tasks, the rectangular window's zero phase distortion and 100\% energy retention make it the best choice for most scenarios.
\begin{table}[h]
\centering

\setlength{\tabcolsep}{3pt}
\renewcommand{\arraystretch}{1.1}

\begin{tabular}{@{}l l |cc|cc|cc@{}}
\toprule
\multirow{2}{*}{\rotatebox[origin=c]{90}{Dataset}} & \multirow{2}{*}{H} & 
\multicolumn{2}{c}{rect} & \multicolumn{2}{c}{hann} & \multicolumn{2}{c@{}}{hamming} \\
\cmidrule(lr){3-4} \cmidrule(lr){5-6} \cmidrule(l){7-8}
 & & MSE & MAE & MSE & MAE & MSE & MAE \\
\midrule

\multirow{5}{*}{\rotatebox[origin=c]{90}{ETTh1}}
& 96 & \firstplace{0.370} & \firstplace{0.392} & 0.373 & 0.394 & 0.370 & 0.392 \\
& 192 & 0.420 & \firstplace{0.419} & \firstplace{0.417} & 0.420 & 0.420 & 0.419 \\
& 336 & 0.460 & 0.440 & \firstplace{0.459} & \firstplace{0.440} & 0.460 & 0.440 \\
& 720 & 0.462 & 0.458 & \firstplace{0.461} & \firstplace{0.457} & 0.462 & 0.458 \\
\cmidrule(lr){2-8}
& AVG & 0.428 & \firstplace{0.427} & \firstplace{0.427} & 0.428 & 0.428 & 0.427 \\
\midrule

\multirow{5}{*}{\rotatebox[origin=c]{90}{ETTh2}}
& 96 & 0.285 & 0.338 & 0.286 & 0.338 & \firstplace{0.285} & \firstplace{0.337} \\
& 192 & \firstplace{0.366} & 0.392 & 0.368 & 0.392 & 0.368 & \firstplace{0.391} \\
& 336 & \firstplace{0.416} & \firstplace{0.431} & 0.420 & 0.433 & 0.419 & 0.433 \\
& 720 & \firstplace{0.435} & \firstplace{0.448} & 0.439 & 0.451 & 0.438 & 0.451 \\
\cmidrule(lr){2-8}
& AVG & \firstplace{0.375} & \firstplace{0.402} & 0.378 & 0.403 & 0.377 & 0.403 \\
\midrule

\multirow{5}{*}{\rotatebox[origin=c]{90}{ETTm1}}
& 96 & \firstplace{0.310} & \firstplace{0.353} & 0.316 & 0.358 & 0.311 & 0.355 \\
& 192 & 0.352 & 0.377 & 0.355 & 0.376 & \firstplace{0.352} & \firstplace{0.375} \\
& 336 & 0.385 & 0.400 & 0.386 & 0.400 & \firstplace{0.385} & \firstplace{0.399} \\
& 720 & 0.448 & 0.438 & 0.449 & 0.437 & \firstplace{0.447} & \firstplace{0.436} \\
\cmidrule(lr){2-8}
& AVG & \firstplace{0.374} & 0.392 & 0.376 & 0.393 & 0.374 & \firstplace{0.391} \\
\midrule

\multirow{5}{*}{\rotatebox[origin=c]{90}{ETTm2}}
& 96 & \firstplace{0.162} & \firstplace{0.244} & 0.164 & 0.247 & 0.163 & 0.245 \\
& 192 & \firstplace{0.227} & \firstplace{0.288} & 0.229 & 0.289 & 0.227 & 0.288 \\
& 336 & \firstplace{0.283} & \firstplace{0.326} & 0.285 & 0.327 & 0.284 & 0.327 \\
& 720 & 0.391 & 0.393 & 0.385 & 0.389 & \firstplace{0.383} & \firstplace{0.387} \\
\cmidrule(lr){2-8}
& AVG & 0.266 & 0.313 & 0.266 & 0.313 & \firstplace{0.264} & \firstplace{0.312} \\
\midrule

\multirow{5}{*}{\rotatebox[origin=c]{90}{Weather}}
& 96 & \firstplace{0.160} & \firstplace{0.205} & 0.164 & 0.208 & 0.162 & 0.206 \\
& 192 & \firstplace{0.208} & \firstplace{0.248} & 0.215 & 0.254 & 0.212 & 0.250 \\
& 336 & \firstplace{0.264} & \firstplace{0.290} & 0.268 & 0.292 & 0.265 & 0.291 \\
& 720 & 0.345 & 0.344 & 0.347 & 0.345 & \firstplace{0.343} & \firstplace{0.343} \\
\cmidrule(lr){2-8}
& AVG & \firstplace{0.244} & \firstplace{0.272} & 0.248 & 0.275 & 0.245 & 0.272 \\
\midrule

\multirow{5}{*}{\rotatebox[origin=c]{90}{ECL}}
& 96 & \firstplace{0.139} & \firstplace{0.233} & 0.147 & 0.243 & 0.141 & 0.235 \\
& 192 & \firstplace{0.155} & \firstplace{0.247} & 0.162 & 0.255 & 0.156 & 0.249 \\
& 336 & \firstplace{0.171} & \firstplace{0.265} & 0.178 & 0.272 & 0.173 & 0.266 \\
& 720 & \firstplace{0.210} & \firstplace{0.297} & 0.216 & 0.303 & 0.211 & 0.299 \\
\cmidrule(lr){2-8}
& AVG & \firstplace{0.169} & \firstplace{0.261} & 0.176 & 0.268 & 0.170 & 0.262 \\
\midrule

\multirow{5}{*}{\rotatebox[origin=c]{90}{Traffic}}
& 96 & \firstplace{0.438} & \firstplace{0.262} & 0.450 & 0.268 & 0.440 & 0.263 \\
& 192 & \firstplace{0.432} & \firstplace{0.260} & 0.445 & 0.266 & 0.436 & 0.262 \\
& 336 & \firstplace{0.443} & \firstplace{0.265} & 0.454 & 0.269 & 0.446 & 0.266 \\
& 720 & \firstplace{0.476} & \firstplace{0.275} & 0.484 & 0.278 & 0.478 & 0.275 \\
\cmidrule(lr){2-8}
& AVG & \firstplace{0.448} & \firstplace{0.265} & 0.458 & 0.270 & 0.450 & 0.266 \\
\bottomrule
\end{tabular}

\caption{Performance comparison of different window functions (rect, hann, hamming) across datasets. The best results are highlighted in red.}
\label{table_windowtype}
\end{table}
\subsubsection{Window Length}
Based on preliminary comparative experiments, we found that setting the window stride \(w_S\) equal to the window length \(w_L\) yields the best performance for a given window length. Therefore, in this comparative study, we consistently maintain \(w_S=w_L\). The window length is selected from \(\{4, 6, 8, 10\}\), and the window type is set as Rectangular Window. Experimental results are presented in Table \ref{table_window_length}.

The experimental results demonstrate that the optimal window length varies across datasets: 4 achieves the best performance on ETTh2, while 6 or 8 yields optimal results on Weather. Therefore, window length should be tuned independently for different datasets through parameter optimization.
\begin{table}[htbp]
\centering

\setlength{\tabcolsep}{2pt}
\renewcommand{\arraystretch}{1.1}

\begin{tabular}{@{}l l |cc|cc|cc|cc@{}}
\toprule
\multirow{2}{*}{\rotatebox[origin=c]{90}{Dataset}} & \multirow{2}{*}{H} & 
\multicolumn{2}{c}{L=4} & \multicolumn{2}{c}{L=6} & \multicolumn{2}{c}{L=8} & \multicolumn{2}{c@{}}{L=10} \\
\cmidrule(lr){3-4} \cmidrule(lr){5-6} \cmidrule(lr){7-8} \cmidrule(l){9-10}
 & & MSE & MAE & MSE & MAE & MSE & MAE & MSE & MAE \\
\midrule

\multirow{5}{*}{\rotatebox[origin=c]{90}{ETTh1}}
& 96 & 0.369 & 0.390 & 0.370 & 0.392 & 0.371 & 0.393 & 0.375 & 0.398 \\
& 192 & 0.417 & 0.418 & 0.420 & 0.419 & 0.418 & 0.418 & 0.421 & 0.423 \\
& 336 & 0.459 & 0.439 & 0.460 & 0.440 & 0.465 & 0.442 & 0.487 & 0.451 \\
& 720 & 0.456 & 0.454 & 0.462 & 0.458 & 0.462 & 0.458 & 0.475 & 0.467 \\
\cmidrule(lr){2-10}
& AVG & 0.425 & 0.425 & 0.428 & 0.427 & 0.429 & 0.428 & 0.440 & 0.435 \\
\midrule

\multirow{5}{*}{\rotatebox[origin=c]{90}{ETTh2}}
& 96 & 0.282 & 0.336 & 0.284 & 0.338 & 0.284 & 0.336 & 0.294 & 0.345 \\
& 192 & 0.361 & 0.388 & 0.362 & 0.390 & 0.370 & 0.397 & 0.368 & 0.393 \\
& 336 & 0.415 & 0.430 & 0.418 & 0.432 & 0.420 & 0.432 & 0.424 & 0.432 \\
& 720 & 0.435 & 0.450 & 0.431 & 0.445 & 0.431 & 0.444 & 0.440 & 0.449 \\
\cmidrule(lr){2-10}
& AVG & 0.373 & 0.401 & 0.374 & 0.401 & 0.376 & 0.402 & 0.381 & 0.405 \\
\midrule

\multirow{5}{*}{\rotatebox[origin=c]{90}{ETTm1}}
& 96 & 0.312 & 0.354 & 0.313 & 0.355 & 0.312 & 0.355 & 0.318 & 0.359 \\
& 192 & 0.352 & 0.374 & 0.351 & 0.375 & 0.351 & 0.375 & 0.359 & 0.381 \\
& 336 & 0.383 & 0.396 & 0.386 & 0.400 & 0.386 & 0.400 & 0.393 & 0.406 \\
& 720 & 0.447 & 0.434 & 0.446 & 0.435 & 0.446 & 0.435 & 0.453 & 0.440 \\
\cmidrule(lr){2-10}
& AVG & 0.373 & 0.390 & 0.374 & 0.391 & 0.374 & 0.391 & 0.381 & 0.396 \\
\midrule

\multirow{5}{*}{\rotatebox[origin=c]{90}{ETTm2}}
& 96 & 0.162 & 0.244 & 0.162 & 0.244 & 0.162 & 0.244 & 0.164 & 0.247 \\
& 192 & 0.226 & 0.287 & 0.227 & 0.288 & 0.226 & 0.286 & 0.228 & 0.290 \\
& 336 & 0.283 & 0.326 & 0.283 & 0.326 & 0.284 & 0.326 & 0.289 & 0.329 \\
& 720 & 0.385 & 0.389 & 0.391 & 0.393 & 0.392 & 0.393 & 0.397 & 0.397 \\
\cmidrule(lr){2-10}
& AVG & 0.264 & 0.311 & 0.266 & 0.313 & 0.266 & 0.313 & 0.269 & 0.316 \\
\midrule

\multirow{5}{*}{\rotatebox[origin=c]{90}{Weather}}
& 96 & 0.161 & 0.204 & 0.160 & 0.205 & 0.159 & 0.204 & 0.170 & 0.217 \\
& 192 & 0.209 & 0.249 & 0.208 & 0.248 & 0.208 & 0.248 & 0.217 & 0.256 \\
& 336 & 0.264 & 0.290 & 0.264 & 0.290 & 0.264 & 0.290 & 0.274 & 0.298 \\
& 720 & 0.344 & 0.344 & 0.345 & 0.344 & 0.345 & 0.344 & 0.350 & 0.348 \\
\cmidrule(lr){2-10}
& AVG & 0.245 & 0.272 & 0.244 & 0.272 & 0.244 & 0.272 & 0.253 & 0.280 \\
\midrule

\multirow{5}{*}{\rotatebox[origin=c]{90}{Electricity}}
& 96 & 0.139 & 0.234 & 0.139 & 0.233 & 0.139 & 0.233 & 0.144 & 0.238 \\
& 192 & 0.155 & 0.248 & 0.155 & 0.247 & 0.155 & 0.247 & 0.158 & 0.251 \\
& 336 & 0.172 & 0.265 & 0.171 & 0.265 & 0.172 & 0.265 & 0.174 & 0.267 \\
& 720 & 0.210 & 0.298 & 0.210 & 0.297 & 0.210 & 0.297 & 0.212 & 0.299 \\
\cmidrule(lr){2-10}
& AVG & 0.169 & 0.261 & 0.169 & 0.261 & 0.169 & 0.261 & 0.172 & 0.264 \\
\midrule

\multirow{5}{*}{\rotatebox[origin=c]{90}{Traffic}}
& 96 & 0.438 & 0.262 & 0.438 & 0.262 & 0.438 & 0.262 & 0.447 & 0.264 \\
& 192 & 0.432 & 0.262 & 0.432 & 0.260 & 0.433 & 0.261 & 0.441 & 0.263 \\
& 336 & 0.443 & 0.265 & 0.443 & 0.265 & 0.444 & 0.264 & 0.449 & 0.266 \\
& 720 & 0.475 & 0.275 & 0.476 & 0.275 & 0.475 & 0.274 & 0.481 & 0.275 \\
\cmidrule(lr){2-10}
& AVG & 0.447 & 0.266 & 0.448 & 0.265 & 0.447 & 0.265 & 0.454 & 0.267 \\
\bottomrule
\end{tabular}

\caption{Performance comparison of different window lengths (L=4,6,8,10) across datasets.}
\label{table_window_length}
\end{table}

\section{D. Model Analysis} 
\subsubsection{Effectiveness of FECF}
Our FECF technique fundamentally operates as a Seasonal-Trend Decomposition (STD) method. To validate its effectiveness, we conducted experiments comparing it against two established STD approaches: the moving average decomposition (MOV) used in DLinear  and the local decomposition (LD) . As demonstrated in Table \ref{table2}, when replacing FECF with either MOV+Linear (DFreq) or LD+Linear (LDFreq) variants in our FreqCycle framework, the forecasting performance consistently degraded across all seven benchmark datasets. These empirical results clearly establish that our FECF-based decomposition achieves superior temporal pattern extraction compared to conventional STD methods.
  
\begin{table}[htbp]
\centering

\setlength{\tabcolsep}{5pt}
\renewcommand{\arraystretch}{1.1}

\begin{tabular}{@{}l l |cc|cc|cc@{}}
\toprule
\multirow{2}{*}{\rotatebox[origin=c]{90}{Dataset}} & \multirow{2}{*}{H} & 
\multicolumn{2}{c}{FreqCycle} & \multicolumn{2}{c}{DFreq} & \multicolumn{2}{c@{}}{LDFreq} \\
\cmidrule(lr){3-4} \cmidrule(lr){5-6} \cmidrule(l){7-8}
 & & MSE & MAE & MSE & MAE & MSE & MAE \\
\midrule

\multirow{5}{*}{\rotatebox[origin=c]{90}{ETTh1}}
& 96 & \firstplace{0.369} & \firstplace{0.390} & 0.385 & 0.402 & 0.385 & 0.402 \\
& 192 & \firstplace{0.417} & \firstplace{0.418} & 0.436 & 0.435 & 0.438 & 0.433 \\
& 336 & \firstplace{0.459} & \firstplace{0.439} & 0.486 & 0.467 & 0.496 & 0.469 \\
& 720 & \firstplace{0.456} & \firstplace{0.454} & 0.559 & 0.537 & 0.555 & 0.526 \\
\cmidrule(lr){2-8}
& AVG. & \firstplace{0.425} & \firstplace{0.425} & 0.467 & 0.460 & 0.468 & 0.457 \\
\midrule

\multirow{5}{*}{\rotatebox[origin=c]{90}{ETTh2}}
& 96 & \firstplace{0.282} & \firstplace{0.336} & 0.323 & 0.378 & 0.304 & 0.362 \\
& 192 & \firstplace{0.361} & \firstplace{0.388} & 0.432 & 0.447 & 0.419 & 0.431 \\
& 336 & \firstplace{0.415} & \firstplace{0.428} & 0.528 & 0.505 & 0.478 & 0.471 \\
& 720 & \firstplace{0.425} & \firstplace{0.443} & 0.737 & 0.616 & 0.680 & 0.591 \\
\cmidrule(lr){2-8}
& AVG. & \firstplace{0.371} & \firstplace{0.399} & 0.505 & 0.487 & 0.470 & 0.464 \\
\midrule

\multirow{5}{*}{\rotatebox[origin=c]{90}{ETTm1}}
& 96 & \firstplace{0.310} & \firstplace{0.353} & 0.334 & 0.372 & 0.346 & 0.378 \\
& 192 & \firstplace{0.351} & \firstplace{0.374} & 0.376 & 0.398 & 0.377 & 0.399 \\
& 336 & \firstplace{0.383} & \firstplace{0.396} & 0.406 & 0.422 & 0.416 & 0.434 \\
& 720 & \firstplace{0.446} & \firstplace{0.434} & 0.464 & 0.458 & 0.468 & 0.456 \\
\cmidrule(lr){2-8}
& AVG. & \firstplace{0.372} & \firstplace{0.389} & 0.395 & 0.413 & 0.402 & 0.417 \\
\midrule

\multirow{5}{*}{\rotatebox[origin=c]{90}{ETTm2}}
& 96 & \firstplace{0.162} & \firstplace{0.244} & 0.188 & 0.277 & 0.186 & 0.274 \\
& 192 & \firstplace{0.226} & \firstplace{0.286} & 0.265 & 0.335 & 0.253 & 0.320 \\
& 336 & \firstplace{0.283} & \firstplace{0.326} & 0.341 & 0.388 & 0.330 & 0.375 \\
& 720 & \firstplace{0.383} & \firstplace{0.387} & 0.457 & 0.461 & 0.548 & 0.503 \\
\cmidrule(lr){2-8}
& AVG. & \firstplace{0.263} & \firstplace{0.311} & 0.312 & 0.365 & 0.329 & 0.368 \\
\midrule

\multirow{5}{*}{\rotatebox[origin=c]{90}{Weather}}
& 96 & \firstplace{0.159} & \firstplace{0.203} & 0.188 & 0.245 & 0.176 & 0.230 \\
& 192 & \firstplace{0.208} & \firstplace{0.248} & 0.235 & 0.297 & 0.216 & 0.266 \\
& 336 & \firstplace{0.264} & \firstplace{0.289} & 0.274 & 0.321 & 0.266 & 0.311 \\
& 720 & 0.343 & \firstplace{0.342} & 0.341 & 0.375 & \firstplace{0.335} & 0.358 \\
\cmidrule(lr){2-8}
& AVG. & \firstplace{0.243} & \firstplace{0.270} & 0.259 & 0.309 & 0.248 & 0.291 \\
\midrule

\multirow{5}{*}{\rotatebox[origin=c]{90}{ECL}}
& 96 & \firstplace{0.140} & \firstplace{0.234} & 0.190 & 0.276 & 0.184 & 0.270 \\
& 192 & \firstplace{0.154} & \firstplace{0.246} & 0.190 & 0.278 & 0.187 & 0.275 \\
& 336 & \firstplace{0.170} & \firstplace{0.264} & 0.203 & 0.294 & 0.198 & 0.288 \\
& 720 & \firstplace{0.210} & \firstplace{0.297} & 0.235 & 0.323 & 0.234 & 0.319 \\
\cmidrule(lr){2-8}
& AVG. & \firstplace{0.168} & \firstplace{0.260} & 0.204 & 0.293 & 0.201 & 0.288 \\
\midrule

\multirow{5}{*}{\rotatebox[origin=c]{90}{Traffic}}
& 96 & \firstplace{0.438} & \firstplace{0.262} & 0.523 & 0.334 & 0.528 & 0.323 \\
& 192 & \firstplace{0.434} & \firstplace{0.258} & 0.497 & 0.310 & 0.501 & 0.301 \\
& 336 & \firstplace{0.446} & \firstplace{0.258} & 0.517 & 0.307 & 0.515 & 0.298 \\
& 720 & \firstplace{0.476} & \firstplace{0.265} & 0.577 & 0.315 & 0.565 & 0.307 \\
\cmidrule(lr){2-8}
& AVG. & \firstplace{0.448} & \firstplace{0.261} & 0.529 & 0.316 & 0.527 & 0.307 \\
\bottomrule
\end{tabular}

\caption{The full result of comparison between STD methods. We set the lookback window size \(L\) as 96 and the prediction length as \(H \in \{96, 192, 336, 720\}\). The best results are in red.}
\label{table2}
\end{table}

\subsubsection{The feed-forward layer in SFPL}
We compared the performance of using Linear layers versus MLP as the feed-forward layers in the SFPL module, with experimental results shown in Table \ref{table3}. The analysis demonstrates that employing MLP as the feed-forward layer achieves superior performance across almost all test datasets. This phenomenon may be attributed to MLP's stronger nonlinear representation capability, enabling it to better capture complex dependencies in time series data.

\begin{table}[htbp]
\centering

\setlength{\tabcolsep}{3pt}
\renewcommand{\arraystretch}{1.1}

\begin{tabular}{@{}l l |cc|cc@{}}
\toprule
\multirow{2}{*}{\rotatebox[origin=c]{90}{Dataset}} & \multirow{2}{*}{H} & 
\multicolumn{2}{c}{FreqCycle(MLP)} & \multicolumn{2}{c@{}}{FreqCycle(Linear)} \\
\cmidrule(lr){3-4} \cmidrule(l){5-6}
 & & \multicolumn{1}{c}{MSE} & \multicolumn{1}{c}{MAE} & \multicolumn{1}{c}{MSE} & \multicolumn{1}{c}{MAE} \\
\midrule

\multirow{5}{*}{\rotatebox[origin=c]{90}{ETTh1}}
& 96 & \firstplace{0.369} & \firstplace{0.390} & 0.370 & 0.392 \\
& 192 & \firstplace{0.417} & \firstplace{0.418} & 0.420 & 0.419 \\
& 336 & \firstplace{0.459} & \firstplace{0.439} & 0.460 & 0.440 \\
& 720 & \firstplace{0.456} & \firstplace{0.454} & 0.462 & 0.458 \\
\cmidrule(lr){2-6}
& AVG & \firstplace{0.425} & \firstplace{0.425} & 0.428 & 0.427 \\
\midrule

\multirow{5}{*}{\rotatebox[origin=c]{90}{ETTh2}}
& 96 & \firstplace{0.282} & \firstplace{0.336} & 0.285 & 0.337 \\
& 192 & \firstplace{0.361} & \firstplace{0.388} & 0.368 & 0.391 \\
& 336 & \firstplace{0.415} & \firstplace{0.428} & 0.419 & 0.433 \\
& 720 & \firstplace{0.425} & \firstplace{0.443} & 0.438 & 0.451 \\
\cmidrule(lr){2-6}
& AVG & \firstplace{0.371} & \firstplace{0.399} & 0.377 & 0.403 \\
\midrule

\multirow{5}{*}{\rotatebox[origin=c]{90}{ETTm1}}
& 96 & \firstplace{0.310} & \firstplace{0.353} & 0.311 & 0.355 \\
& 192 & \firstplace{0.351} & \firstplace{0.374} & 0.352 & 0.375 \\
& 336 & \firstplace{0.383} & \firstplace{0.396} & 0.385 & 0.399 \\
& 720 & \firstplace{0.446} & \firstplace{0.434} & 0.447 & 0.436 \\
\cmidrule(lr){2-6}
& AVG & \firstplace{0.372} & \firstplace{0.389} & 0.374 & 0.391 \\
\midrule

\multirow{5}{*}{\rotatebox[origin=c]{90}{ETTm2}}
& 96 & \firstplace{0.162} & \firstplace{0.244} & 0.163 & 0.245 \\
& 192 & \firstplace{0.226} & \firstplace{0.286} & 0.227 & 0.288 \\
& 336 & \firstplace{0.283} & \firstplace{0.326} & 0.284 & 0.327 \\
& 720 & \firstplace{0.383} & \firstplace{0.387} & 0.383 & 0.387 \\
\cmidrule(lr){2-6}
& AVG & \firstplace{0.263} & \firstplace{0.311} & 0.264 & 0.312 \\
\midrule

\multirow{5}{*}{\rotatebox[origin=c]{90}{Weather}}
& 96 & \firstplace{0.159} & \firstplace{0.203} & 0.162 & 0.206 \\
& 192 & \firstplace{0.208} & \firstplace{0.248} & 0.212 & 0.250 \\
& 336 & \firstplace{0.264} & \firstplace{0.289} & 0.265 & 0.291 \\
& 720 & \firstplace{0.343} & \firstplace{0.342} & 0.343 & 0.343 \\
\cmidrule(lr){2-6}
& AVG & \firstplace{0.243} & \firstplace{0.270} & 0.245 & 0.272 \\
\midrule

\multirow{5}{*}{\rotatebox[origin=c]{90}{ECL}}
& 96 & \firstplace{0.140} & \firstplace{0.234} & 0.141 & 0.235 \\
& 192 & \firstplace{0.154} & \firstplace{0.246} & 0.156 & 0.249 \\
& 336 & \firstplace{0.170} & \firstplace{0.264} & 0.173 & 0.266 \\
& 720 & \firstplace{0.210} & \firstplace{0.297} & 0.211 & 0.299 \\
\cmidrule(lr){2-6}
& AVG & \firstplace{0.168} & \firstplace{0.260} & 0.170 & 0.262 \\
\midrule

\multirow{5}{*}{\rotatebox[origin=c]{90}{Traffic}}
& 96 & \firstplace{0.438} & \firstplace{0.262} & 0.440 & 0.263 \\
& 192 & \firstplace{0.434} & \firstplace{0.258} & 0.436 & 0.262 \\
& 336 & \firstplace{0.446} & \firstplace{0.258} & 0.446 & 0.266 \\
& 720 & \firstplace{0.476} & \firstplace{0.265} & 0.478 & 0.275 \\
\cmidrule(lr){2-6}
& AVG & \firstplace{0.448} & \firstplace{0.261} & 0.450 & 0.266 \\
\bottomrule
\end{tabular}

\caption{Comparison between FreqCycle(MLP) and FreqCycle(Linear) models across different datasets and prediction horizons. The best results are highlighted in red.}
\label{table3}
\end{table}

\subsubsection{Full results of MFreqCycle}
To evaluate the effectiveness of the MFreqCycle model, we design Table \ref{table4} with lookback windows of 96 and 672(168), comparing the model against CycleNet and FITS. As shown in Table \ref{table4}, simply increasing the lookback window may lead to minor performance improvements or even slight degradation in some cases. The reason could be that the inherent limitations of the models prevent them from handling the complex temporal dependencies or noise interference introduced by longer lookback windows. However, the MFreqCycle model, specifically designed for longer lookback windows, achieves remarkable performance improvements on the ETT datasets, demonstrating its pioneering capability in learning both long-period patterns and their aperiodic components.

\begin{table*}[htbp]
\centering
\setlength{\tabcolsep}{3pt}
\renewcommand{\arraystretch}{1.1}

\begin{tabular}{@{}l l |cc|cc|cc|cc|cc|cc@{}}
\toprule
\multirow{2}{*}{\rotatebox[origin=c]{90}{Dataset}} & \multirow{2}{*}{H} & 
\multicolumn{2}{c}{FreqCycle L=96} & \multicolumn{2}{c}{MFreqCycle L=672/168} & \multicolumn{2}{c}{CycleNet L=96} & \multicolumn{2}{c}{CycleNet L=672/168} & \multicolumn{2}{c}{FITS L=96} & \multicolumn{2}{c@{}}{FITS L=672/168} \\
\cmidrule(lr){3-4} \cmidrule(lr){5-6} \cmidrule(lr){7-8} \cmidrule(lr){9-10} \cmidrule(lr){11-12} \cmidrule(l){13-14}
 & & MSE & MAE & MSE & MAE & MSE & MAE & MSE & MAE & MSE & MAE & MSE & MAE \\
\midrule

\multirow{5}{*}{\rotatebox[origin=c]{90}{ETTh2}}
& 96 & 0.282 & 0.336 & \firstplace{0.183} & \firstplace{0.296} & 0.285 & 0.335 & 0.307 & 0.360 & 0.295 & 0.350 & 0.271 & 0.336 \\
& 192 & 0.361 & 0.388 & \firstplace{0.215} & \firstplace{0.322} & 0.373 & 0.391 & 0.385 & 0.406 & 0.381 & 0.396 & 0.331 & 0.374 \\
& 336 & 0.415 & 0.428 & \firstplace{0.251} & \firstplace{0.349} & 0.421 & 0.433 & 0.423 & 0.440 & 0.426 & 0.438 & 0.354 & 0.395 \\
& 720 & 0.425 & 0.443 & \firstplace{0.313} & \firstplace{0.395} & 0.453 & 0.458 & 0.436 & 0.454 & 0.431 & 0.446 & 0.377 & 0.422 \\
\cmidrule(lr){2-14}
& AVG & 0.371 & 0.399 & \firstplace{0.241} & \firstplace{0.340} & 0.383 & 0.404 & 0.388 & 0.415 & 0.383 & 0.408 & 0.333 & 0.382 \\
\midrule

\multirow{5}{*}{\rotatebox[origin=c]{90}{ETTm2}}
& 96 & 0.162 & 0.244 & \firstplace{0.123} & \firstplace{0.239} & 0.166 & 0.248 & 0.176 & 0.269 & 0.183 & 0.266 & 0.163 & 0.254 \\
& 192 & 0.226 & 0.286 & \firstplace{0.146} & \firstplace{0.262} & 0.233 & 0.291 & 0.231 & 0.305 & 0.247 & 0.305 & 0.217 & 0.291 \\
& 336 & 0.283 & 0.326 & \firstplace{0.178} & \firstplace{0.287} & 0.293 & 0.330 & 0.282 & 0.339 & 0.307 & 0.342 & 0.269 & 0.326 \\
& 720 & 0.383 & 0.387 & \firstplace{0.222} & \firstplace{0.326} & 0.395 & 0.389 & 0.357 & 0.388 & 0.407 & 0.399 & 0.351 & 0.379 \\
\cmidrule(lr){2-14}
& AVG & 0.263 & 0.311 & \firstplace{0.167} & \firstplace{0.279} & 0.272 & 0.315 & 0.262 & 0.325 & 0.286 & 0.328 & 0.250 & 0.312 \\
\bottomrule
\end{tabular}

\caption{Performance comparison of different models with varying lookback window lengths (L=96 vs L=672/168). The best results are highlighted in red. Results are averaged across all prediction horizons. }
\label{table4}
\end{table*}

\subsection{E. Visualization}
\subsubsection{Visualization of SFPL's Key Frequency Enhancement}
\begin{figure*}[h]
\centering
\includegraphics[width=2.1\columnwidth]{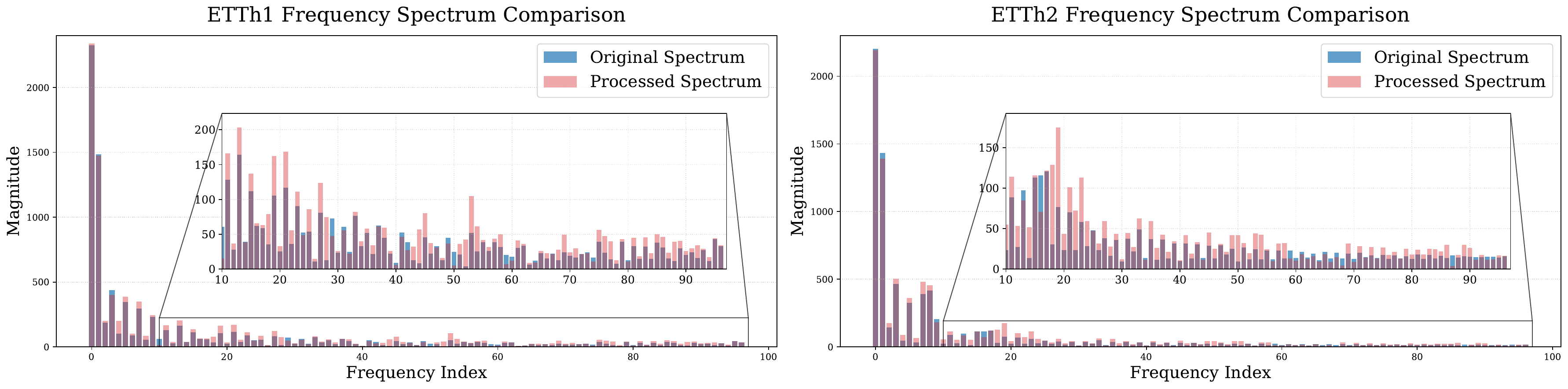}
\caption{Frequency spectrum of ETTh datasets before and
after SFPL enhancement}
\label{freq_ETTh}
\end{figure*}

\begin{figure*}[h]
\centering
\includegraphics[width=2.1\columnwidth]{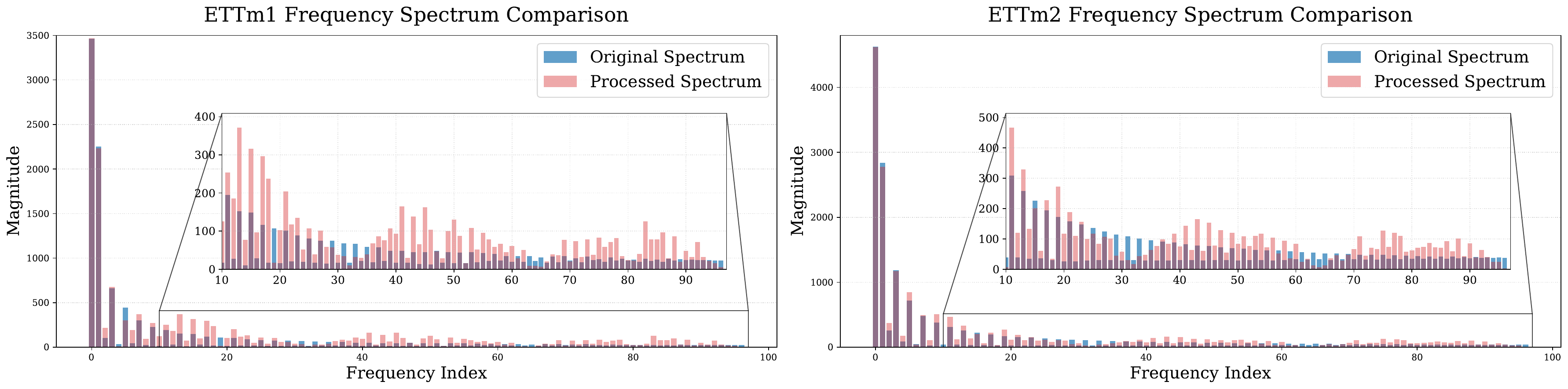}
\caption{Frequency spectrum of ETTm datasets before and
after SFPL enhancement}
\label{freq_ETTm}
\end{figure*}

\begin{figure*}[h]
\centering
\includegraphics[width=2.1\columnwidth]{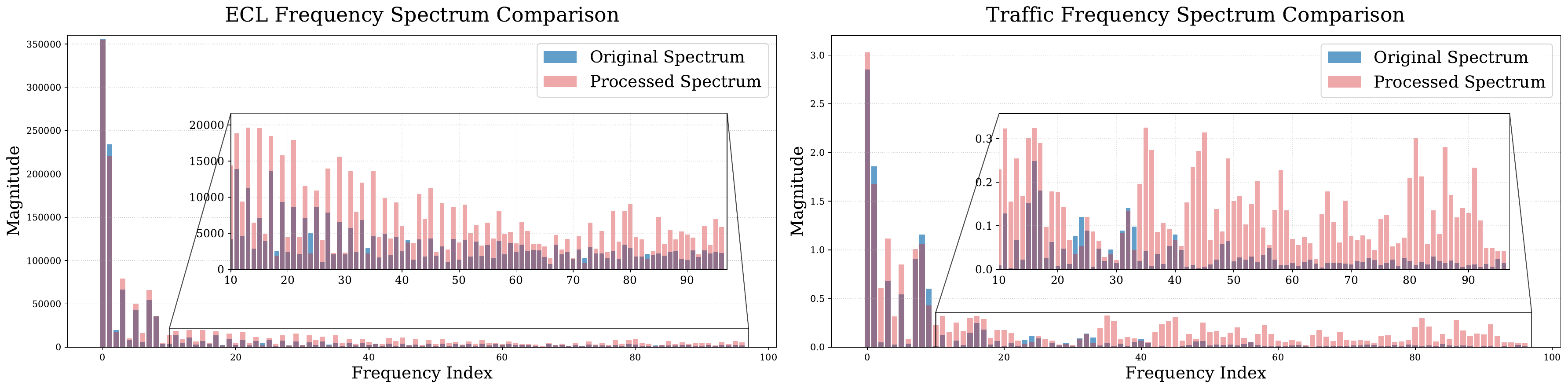}
\caption{Frequency spectrum of Electricity and Traffic datasets before and
after SFPL enhancement}
\label{freq_ET}
\end{figure*}

\begin{figure*}[h]
\centering
\includegraphics[width=1.2\columnwidth]{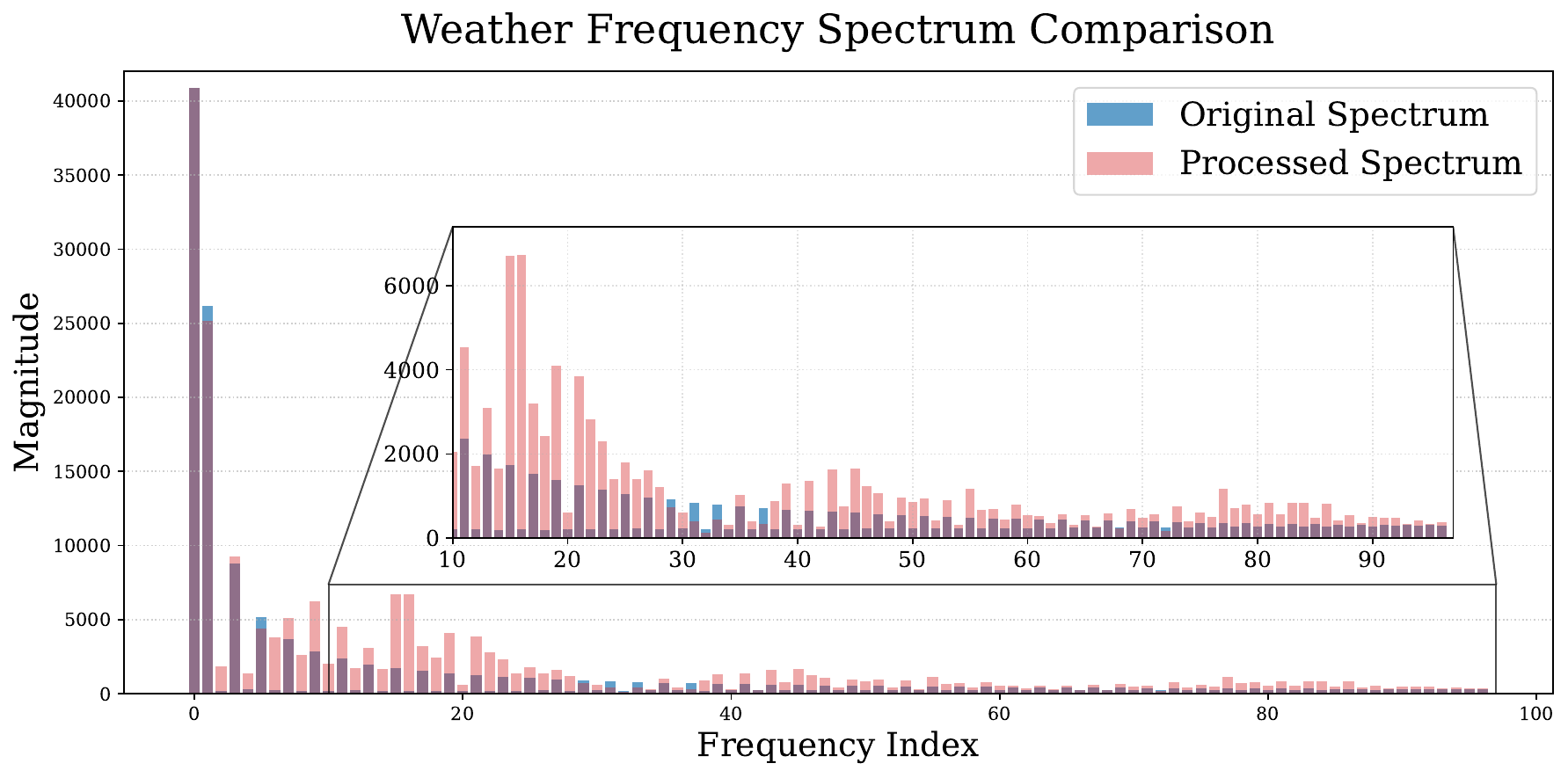}
\caption{Frequency spectrum of Weather dataset before and
after SFPL enhancement}
\label{freq_We}
\end{figure*}

After training the model on benchmark datasets, we analyzed the frequency amplitude spectra before and after SFPL processing. As demonstrated in Figure \ref{freq_ETTh}, \ref{freq_ETTm}, \ref{freq_ET}, \ref{freq_We} SFPL effectively Amplifies the energy proportion of mid to high frequency bands as well as Preserves the structural integrity of dominant low-frequency components.

\subsubsection{Visualization of Forecasting Results}
\begin{figure*}
\centering
\includegraphics[width=2\columnwidth]{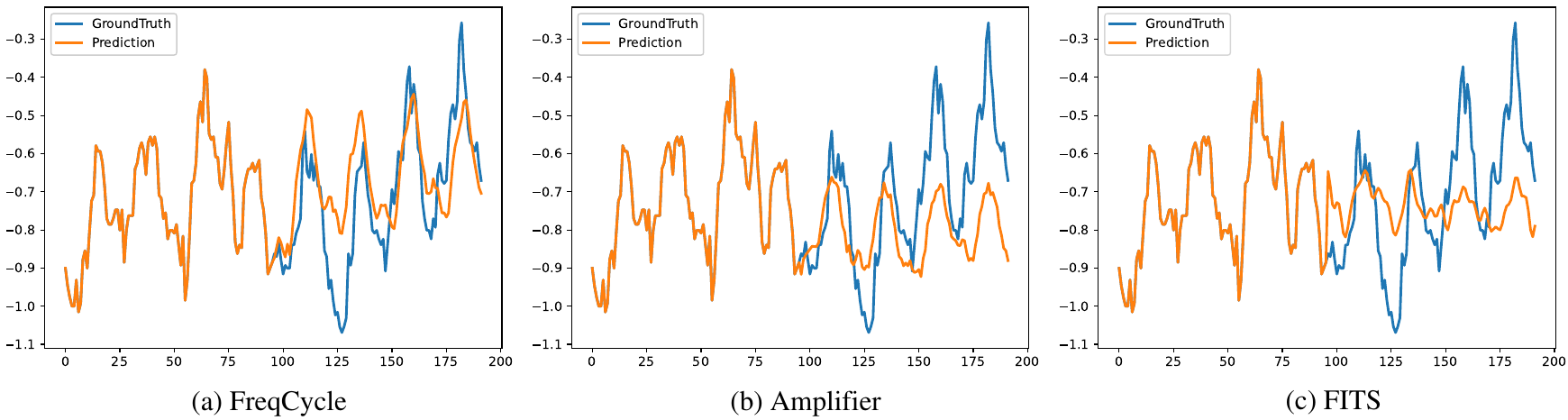} 
\caption{The prediction results of ETTh1. With L=96 and H=96. }
\label{figETTh1}
\end{figure*}

\begin{figure*}
\centering
\includegraphics[width=2\columnwidth]{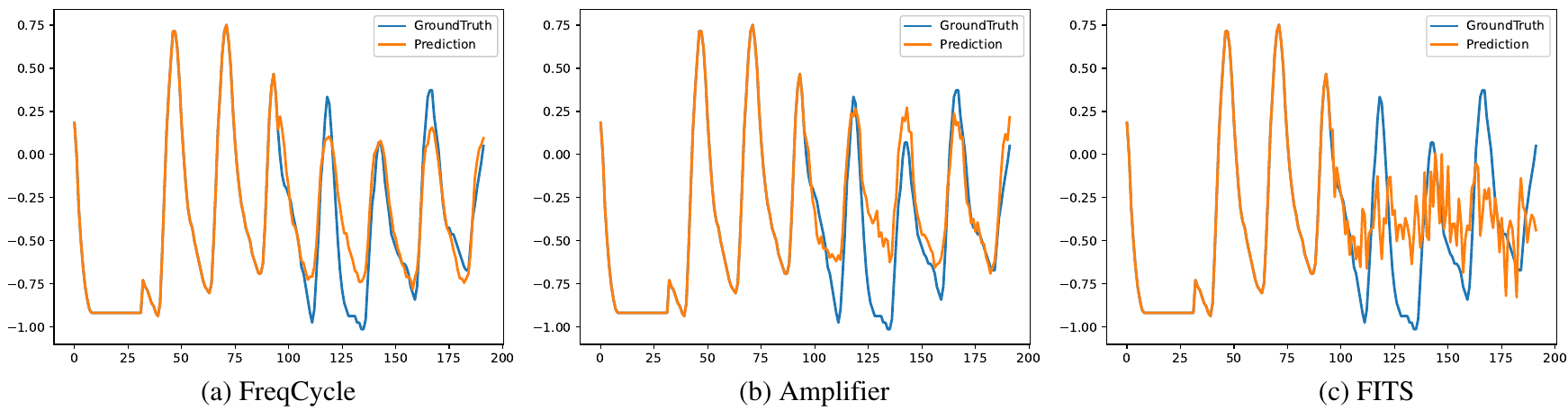} 
\caption{The prediction results of ETTh2. With L=96 and H=96. }
\label{figETTh2}
\end{figure*}

\begin{figure*}
\centering
\includegraphics[width=2\columnwidth]{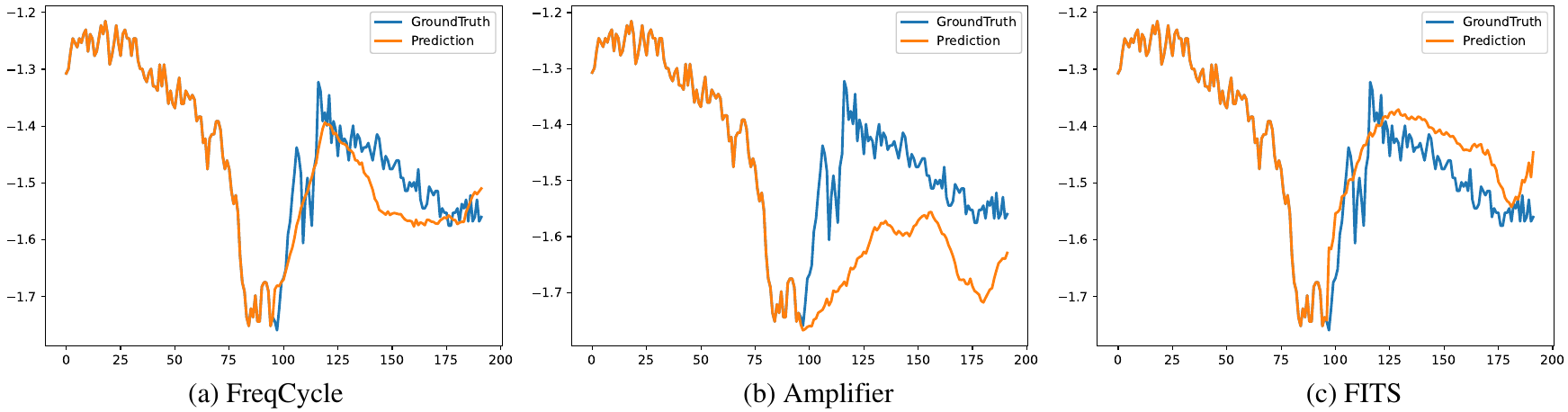} 
\caption{The prediction results of ETTm1. With L=96 and H=96. }
\label{figETTm1}
\end{figure*}

\begin{figure*}
\centering
\includegraphics[width=2\columnwidth]{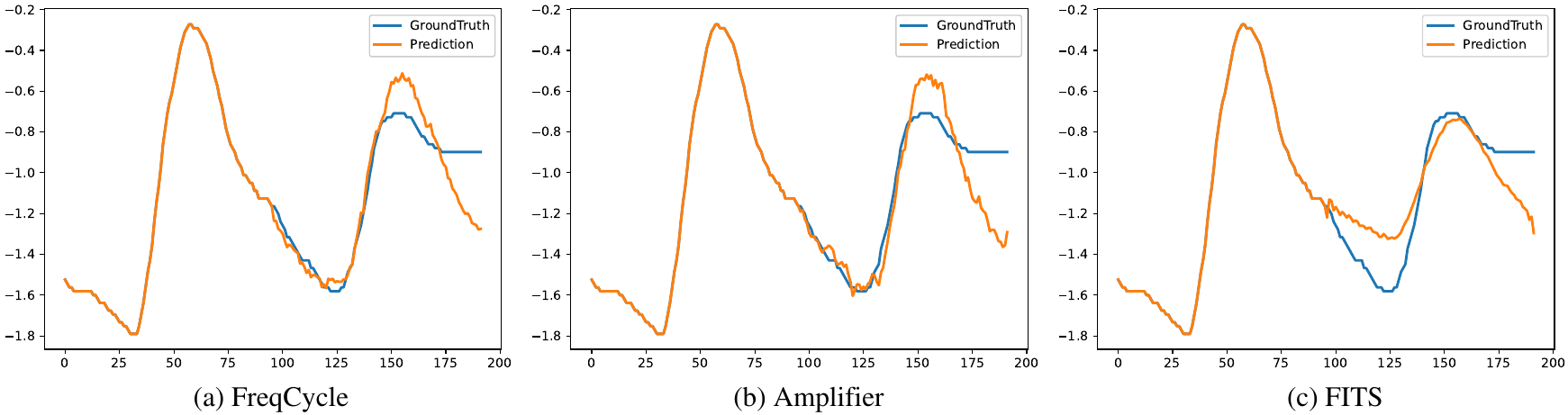} 
\caption{The prediction results of ETTm2. With L=96 and H=96. }
\label{figETTm2}
\end{figure*}

\begin{figure*}
\centering
\includegraphics[width=2\columnwidth]{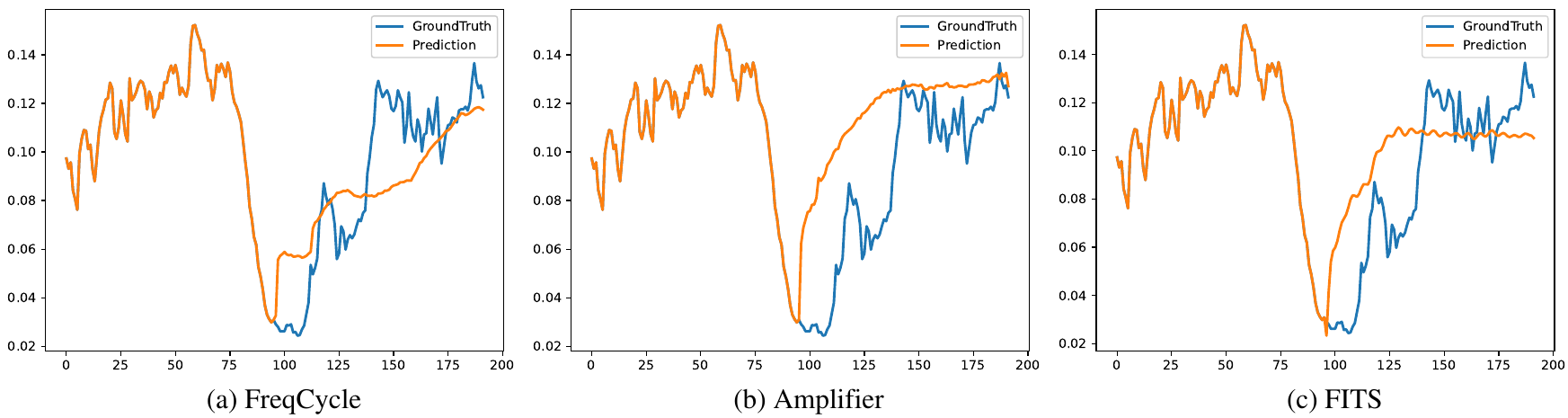} 
\caption{The prediction results of Weather. With L=96 and H=96. }
\label{figWeather}
\end{figure*}

\begin{figure*}
\centering
\includegraphics[width=2\columnwidth]{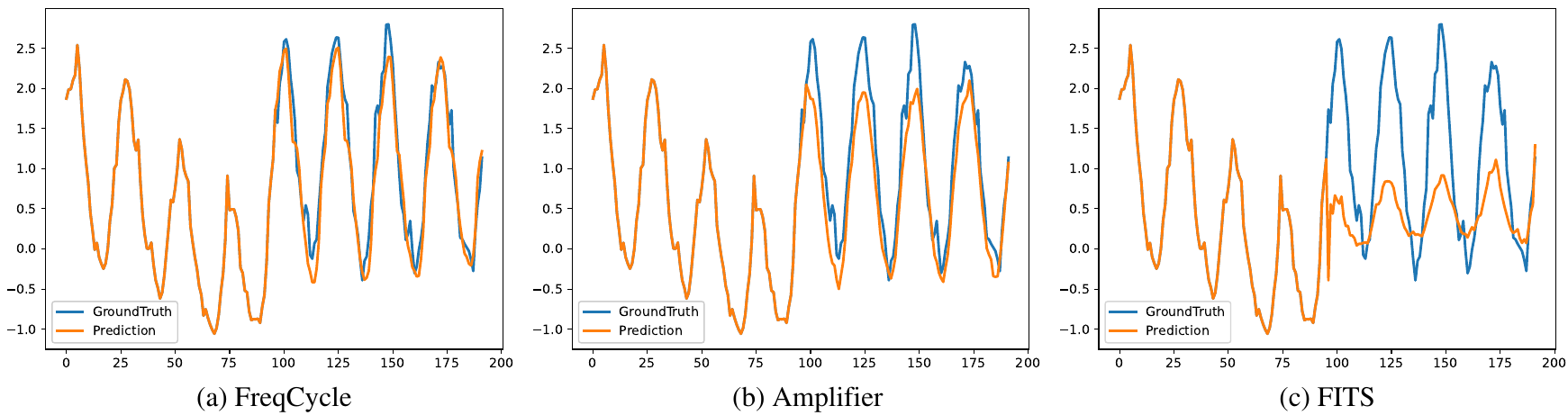} 
\caption{The prediction results of Electricity. With L=96 and H=96. }
\label{figElectricity}
\end{figure*}

\begin{figure*}
\centering
\includegraphics[width=2\columnwidth]{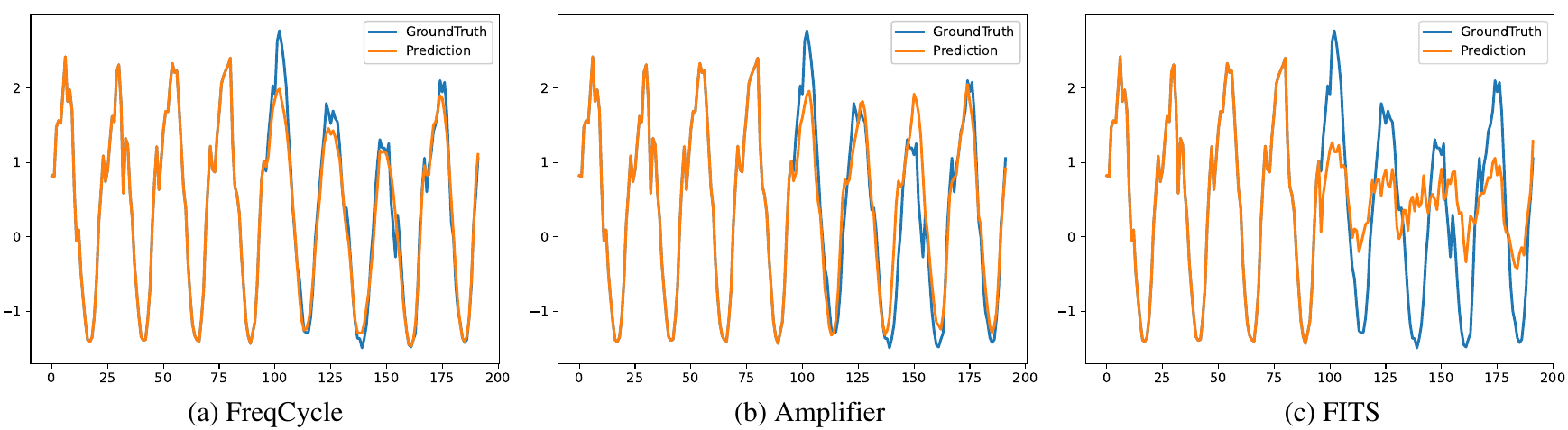} 
\caption{The prediction results of Traffic. With L=96 and H=96. }
\label{figtraffic}
\end{figure*}

The prediction results on benchmark datasets are shown in Figure \ref{figETTh1},\ref{figETTh2},\ref{figETTm1},\ref{figETTm2},\ref{figWeather},\ref{figElectricity},\ref{figtraffic}.


\end{document}